\documentclass[10pt,twocolumn,letterpaper]{article}

\usepackage[pagenumbers]{iccv} 

\usepackage{multirow}
\usepackage{enumitem}
\usepackage{makecell}
\usepackage{amsmath}
\usepackage{amssymb}
\usepackage{graphicx}
\usepackage{subcaption}
\usepackage{gensymb}
\usepackage{comment}
\usepackage{diagbox} 
\usepackage{booktabs}

\newcommand{\Tref}[1]{Table~\ref{#1}}

\newcommand{\Eref}[1]{Eq.~(\ref{#1})}

\newcommand{\Fref}[1]{Figure~\ref{#1}}
\newcommand{\sref}[1]{Sec.~\ref{#1}}

\newcommand{\textblock}[1]{\vspace{3pt}\noindent\textbf{#1}\hspace{0.2em}}

\def\eg{\emph{e.g.}} 
\def\ie{\emph{i.e.}}

\def\ourPaperTitle {Representing 3D Shapes with 64 Latent Vectors for 3D Diffusion Models}
\def\ourPaperId {3586} 

\definecolor{iccvblue}{rgb}{0.21,0.49,0.74}
\usepackage[pagebackref,breaklinks,colorlinks,allcolors=iccvblue]{hyperref}

\def\paperID{\ourPaperId} 
\def\confName{ICCV}
\def\confYear{2025}

\title{\ourPaperTitle}

\author{In Cho \quad Youngbeom Yoo \quad Subin Jeon \quad Seon Joo Kim\\[1mm]
Yonsei University\\
{\tt\small\{join, youngbeom.yoo, subinjeon, seonjookim\}@yonsei.ac.kr}
}

\begin{document}
\maketitle

\begin{abstract}
Constructing a compressed latent space through a variational autoencoder (VAE) is the key for efficient 3D diffusion models.
This paper introduces COD-VAE that encodes 3D shapes into a COmpact set of 1D latent vectors without sacrificing quality.
COD-VAE introduces a two-stage autoencoder scheme to improve compression and decoding efficiency.
First, our encoder block progressively compresses point clouds into compact latent vectors via intermediate point patches. Second, our triplane-based decoder reconstructs dense triplanes from latent vectors instead of directly decoding neural fields, significantly reducing computational overhead of neural fields decoding. Finally, we propose uncertainty-guided token pruning, which allocates resources adaptively by skipping computations in simpler regions and improves the decoder efficiency.
Experimental results demonstrate that COD-VAE achieves 16$\times$ compression compared to the baseline while maintaining quality. This enables $20.8\times$ speedup in generation, highlighting that a large number of latent vectors is not a prerequisite for high-quality reconstruction and generation.
The code is available at \href{https://github.com/join16/COD-VAE}{https://github.com/join16/COD-VAE}.
\end{abstract}
\section{Introduction}

Learning a compressed latent space through a variational autoencoder \cite{kingma2014vae} (VAE) has become a crucial component of recent diffusion models. Due to their iterative inference procedures, generating contents directly in the observation space (\eg, pixels) is computationally intractable.
To address this challenge, most state-of-the-art generative models adopt a latent diffusion framework: a VAE first compresses the input data into a lower-dimensional latent space, and diffusion models are trained within this latent space.

Constructing a compact, well-structured latent space is even more crucial in the 3D domain.
3D objects are inherently irregular, sparse and continuous, which makes direct generation in the 3D domain more challenging.
Previous research has explored various explicit 3D representations, \eg, points \cite{vahdat2022lion}, voxels \cite{ren2024xcube}, and octrees \cite{xiong2024octfusion}.
These explicit representations typically demand larger latent sizes and specifically designed networks to process, making the training and the scaling of generative models difficult.

\begin{figure}[t]
    \centering
    \includegraphics[width=1\linewidth]{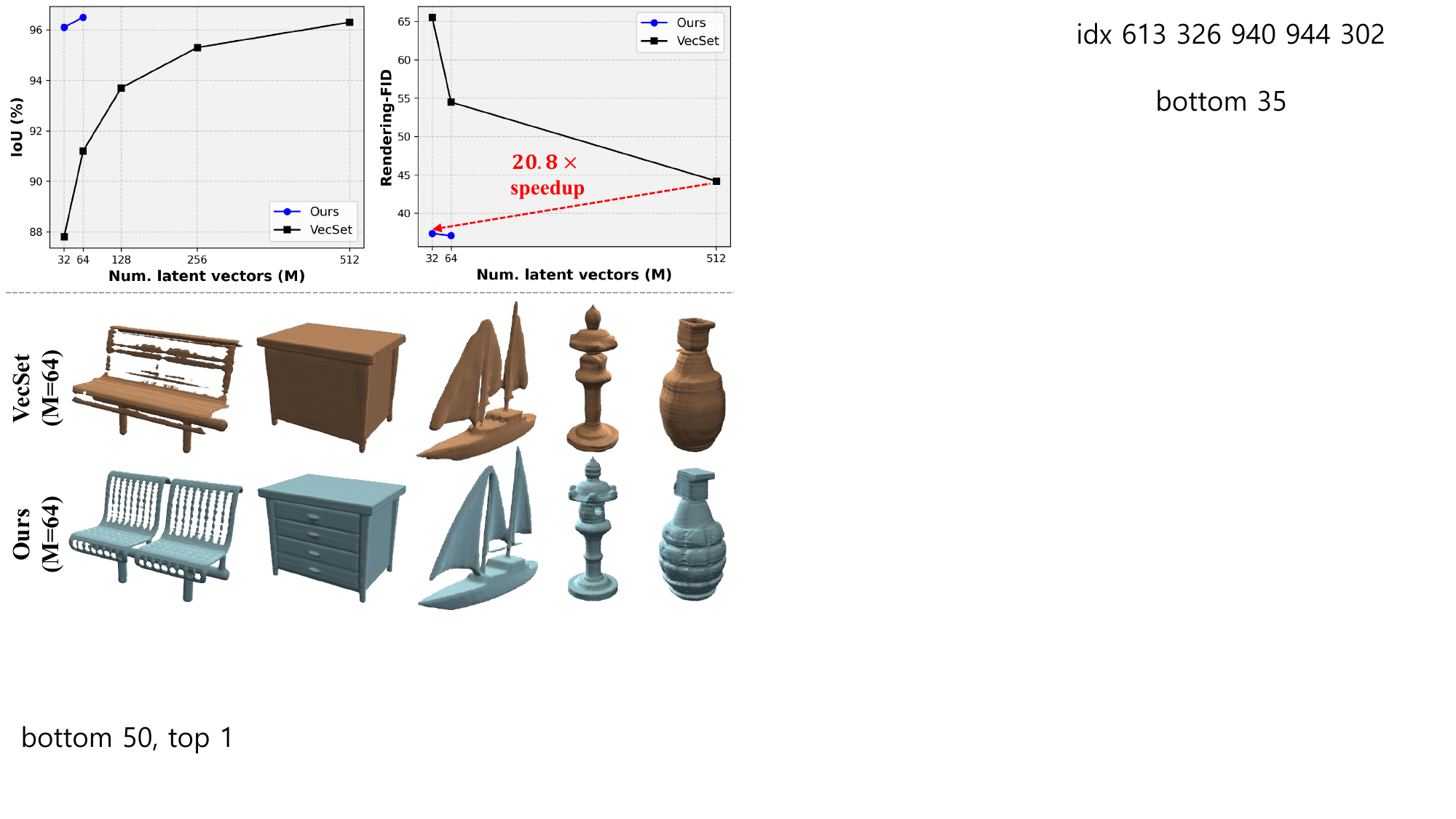}
    \vspace{-15pt}
    \caption{
    \textbf{(top)} Reconstruction IoU (left) and Rendering-FID of the generation results (right) with varying numbers of latent vectors ($M$). Our COD-VAE outperforms VecSet using 16$\times$ fewer latent vectors, achieving $20.8\times$ generation speedup.
    \textbf{(bottom)} VecSet with $M=64$ struggles to capture details, while our model accurately reconstructs detailed and complex shapes of the objects.
    }
    \label{fig:teaser}
    \vspace{-10pt}
\end{figure}
Recently, 3DShape2VecSet \cite{zhang2023vecset} (VecSet) established a foundational framework for encoding 3D shapes into a set of 1D latent vectors.
These 1D vectors provide a more compact latent space than the explicit representations, and are highly compatible with modern transformer implementations (\eg, FlashAttention \cite{dao2022flashattention}).
Due to their simplicity and effectiveness, 1D latent vectors have been actively extended in subsequent research \cite{zhang2024clay, petrov2024gem3d, zhao2024michelangelo, cao2024motion2vecsets, dong2024gpld3d, zhang2024lagem, chen2024dora, zhao2025hunyuan3d}.

The core component of VecSet is the cross-attention layer, which directly projects an input point cloud into 1D latent vectors and maps them back to continuous neural fields.
These direct mappings, operating as learnable downsampling and interpolation, leads to the latent vectors only with a moderate compression ratio.
Due to the limited compression ratio, VecSet still requires sufficiently many latent vectors for high-quality reconstruction, resulting in excessive computational costs of 3D generative models.
Moreover, directly mapping latent vectors to neural fields through cross-attention creates another computational bottleneck.

In this paper, we introduce COD-VAE, a VAE that encodes 3D shapes into a \underline{CO}mpact set of 1\underline{D} vectors with an improved compression ratio.
COD-VAE replaces the direct mappings between points and latent vectors with a two-stage autoencoder scheme.
This scheme enables our model to construct a significantly compressed latent space, thereby accelerating subsequent diffusion models.
It achieves high-quality reconstruction with $16\times$ fewer latent vectors, as well as efficiency improvement in neural fields decoding.

The two-stage autoencoder scheme leverages intermediate representation spaces with a moderate compression ratio.
These intermediate representations serve as bridges between 3D points and latent vectors.
Concretely, our COD-VAE incorporates the following core components:
\begin{itemize}
\item 
Our encoder introduces intermediate point patches, which are obtained by leveraging the attention-based downsampling of VecSet as a learnable point patchifier.
Each encoder block progressively transforms high-resolution points into compact latent vectors by first mapping them into intermediate point patches and then further compressing them into latent vectors.
The global information of the latent vectors are mapped back to points at the end of the block, further refining high-resolution features.

\item 
Our decoder leverages triplanes \cite{chan2022eg3d, fridovich2023kplanes} as intermediate representations for efficient yet effective decoding process.
Inspired by \cite{yu2024titok}, we treat dense triplane embeddings as mask tokens, and reconstruct them from the latent vectors using transformers.
This design significantly reduces computational overheads of the neural fields decoding while also improving the reconstruction quality.

\item 
Apart from the efficient neural fields decoding, processing dense triplanes through transformers requires excessive computational resources.
To mitigate this, we propose uncertainty-guided token pruning that reduces computations in simple regions, thereby achieving further efficiency improvement.
Our decoder employs an auxiliary uncertainty head at the beginning, trained to predict reconstruction errors.
By pruning regions with lower uncertainty, our decoder prioritizes computational resources for reconstructing more complex regions.

\end{itemize}

The compact latent space constructed by COD-VAE enables efficient and high-quality 3D generation.
Experimental results on ShapeNet \cite{chang2015shapenet}, and Objaverse \cite{deitke2023objaverse} verify the effectiveness of our method, achieving $20.8\times$ speedup in generation without sacrificing performance (see \Fref{fig:teaser}).
These results highlight that a large number of latent vectors is not a prerequisite for high-quality reconstruction and generation.
We only need 64 latent vectors to surpass the results of VecSet with 512, or even 1024 latent vectors.
\section{Related work}

\textblock{3D generative models.}
Previous methods for 3D generation can be categorized into two main approaches.
The first approach employs an optimization-based pipeline with 2D diffusion priors to generate 3D objects \cite{poole2022dreamfusion, liu2023zero123, lin2023magic3d, shi2023mvdream, tang2023make, qiu2024richdreamer}.
These methods incur extensive computational costs due to the optimization process per object.
Conversely, the second approach directly trains generative models on 3D datasets, which have shown promising results with significantly faster inference times.
These native 3D generative models have been widely explored, including GAN-based models \cite{achlioptas2018learning, shu20193dgan, ibing2021implicitgan, chan2022eg3d, gao2022get3d, niemeyer2021giraffe}, autoregressive models \cite{sun2020pointgrow, nash2020polygen, mittal2022autosdf, yan2022shapeformer}, and normalizing flows \cite{yang2019pointflow, klokov2020discrete, kim2020softflow}.
Recent breakthroughs of diffusion models have also inspired a wave of research in the 3D domain \cite{nichol2022pointe, zheng2023las, shim2023sdfdiffusion, li2023diffusionsdf, jun2023shap, mo2023dit3d}, which typically adopt the latent diffusion framework \cite{nam20223dldm}.

\textblock{3D VAEs.}
Several studies have explored 3D VAEs with various latent representations.
One line of works employs dense structures to model the latent space \cite{cheng2023sdfusion, lan2024ln3diff, liu2024direct}, which lead to quadratic \cite{lan2024ln3diff, liu2024direct} or cubic \cite{cheng2023sdfusion} computational costs as the resolution increases.
Other works have designed a more efficient latent space by leveraging sparse structures, such as points \cite{vahdat2022lion}, octrees \cite{xiong2024octfusion}, sparse voxels \cite{ren2024xcube}, or irregular latent grids \cite{zhang20223dilg}.
However, processing sparse structures is challenging and often requires specialized architectures and generation process.
This makes the training of generative models difficult, limiting the scalability of these methods.

\textblock{1D latent vectors.}
VecSet \cite{zhang2023vecset} proposes a framework to model the latent space through 1D vectors.
Its simplicity, compactness, and effectiveness has inspired a wide range of research \cite{zhang2024clay, dong2024gpld3d, petrov2024gem3d, zhao2024michelangelo, zhang2024lagem, chen2024dora, zhao2025hunyuan3d}.
The major challenge present in this line of works is the number of latent vectors required for high-quality reconstruction.
As the direct mappings through the cross-attention lacks the capability to obtain higher compression ratio, successive generative models suffer from huge computation costs to generate these large sets of latent vectors.
Additionally, decoding neural fields through the direct mapping causes another excessive computation costs.
Our COD-VAE addresses these challenges with a two-stage autoencoder scheme, in which we utilize intermediate representation spaces as bridges.

\textblock{Compact latent space.}
The high computational costs of self-attention has led to efforts in the 2D domain for enhancing latent space compression \cite{yu2024titok, chen2024dcae}.
TiTok \cite{yu2024titok} proposes to represent images with highly compressed 1D latent tokens, which are obtained using transformers.
DC-AE \cite{chen2024dcae} achieves highly compressed 2D latent space to accelerate diffusion models.
We aim to extend this paradigm to the 3D domain, by leveraging learnable point patches and triplanes as intermediate representations.
Concurrent to our work, other studies \cite{gao2025mars, zhang2024vat} propose compact variational tokenizers with next scale prediction for autogressive generation.


\begin{figure*}[t]
    \centering
    \includegraphics[width=1\linewidth]{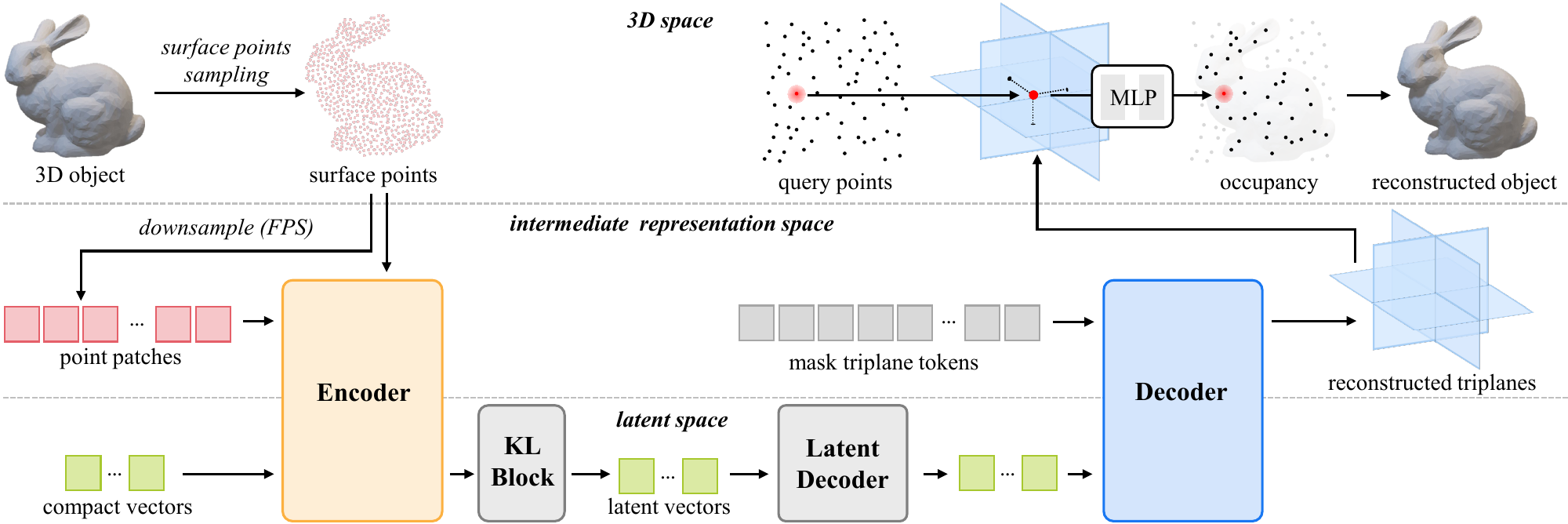}
    \vspace{-15pt}
    \caption{
    \textbf{Overview of COD-VAE.} We propose a two-stage autoencoder scheme to obtain compact 1D latent vectors.
    The encoder leverages intermediate point patches with a moderate compression ratio, and the decoder reconstructs triplanes from the latent vectors.
    }
    \label{fig:method_overview}
    \vspace{-10pt}
\end{figure*}

\section{Method}

\subsection{Preliminary of VecSet}
\label{sec:preliminary}
VecSet \cite{zhang2023vecset} is a transformer-based VAE that directly maps points clouds into a set of 1D latent vectors and decodes them into continuous occupancy values.
Given an input point cloud $\mathcal{P} = \{\mathbf{p}_i\}_{i=1}^{N}$ sampled from the object's surface,
VecSet encodes the point cloud into a set of 1D feature vectors $\mathcal{F} = \{ \mathbf{f}_i \in \mathbb{R}^C \}_{i=1}^M$ using the cross-attention:
\begin{equation}
    \mathcal{F} = \mathrm{CrossAttn}(\gamma(\hat{\mathcal{P}}),\ \gamma(\mathcal{P})),
\end{equation}
where $\gamma(\cdot)$ is a learnable positional embedding, and $\hat{\mathcal{P}}$ is the initial positions of $\mathcal{F}$, obtained by downsampling $\mathcal{P}$ with the farthest-point-sampling (FPS).
Then the 1D latent vectors $\mathcal{Z} = \{\mathbf{z}_i \in \mathbb{R}^D \}_{i=1}^M$, where $D \ll C$, are obtained by compressing $\mathcal{F}$ along the channel dimension:
\begin{equation}
\label{eq:vecset_kl}
    \mathbf{z}_i = \mathrm{FC}_{\mu}(\mathbf{f}_i) + \mathrm{FC}_{\sigma}(\mathbf{f}_i) \cdot \epsilon,
\end{equation}
where $\mathrm{FC}_\mu(\cdot)$, $\mathrm{FC}_\sigma(\cdot)$ are linear layers that project the features into $D$-dimensional mean and variance, and $\epsilon \sim N(0, I)$.
The resulting $\mathcal{Z}$ is a set of multivariate 1D Gaussian features, which is also regularized by KL divergence.

The decoder of VecSet first projects $\mathcal{Z}$ into $C$-dimensional features $\hat{\mathcal{F}}$, and processes them with several self-attention layers.
Finally, a cross-attention layer maps these vectors into continuous occupancy values.
For a query point $\textbf{q} \in \mathbb{R}^3$, the decoding process can be described as
\begin{equation}
\label{eq:vecset_occ}
    o(\mathbf{q}) = \mathrm{FC}_{o}(\mathrm{CrossAttn}(\gamma(\mathbf{q})), \mathrm{SelfAttn}^{L}(\hat{\mathcal{F}})),
\end{equation}
where $\mathrm{SelfAttn}^{L}(\cdot)$ is a stack of $L$ self-attention layers, and $\mathrm{FC}_o: \mathbb{R}^C \rightarrow \mathbb{R}$ is a linear layer.

The cross-attention layers, which directly map points to latent vectors and vice versa, form the core of the VecSet framework.
However, due to their limited compression capability, VecSet struggles to achieve latent vectors over a certain compression ratio.
As a result, VecSet-based methods need a large number of latent vectors to obtain high-quality results, which yields substantial computational costs of diffusion models.
Furthermore, directly decoding latent vectors into neural fields introduces an additional bottleneck, as neural fields decoding involves processing of enormous query points (\ie, over 2M points for $128^3$ grids).

\subsection{COD-VAE architecture}
\label{sec:ours}
We introduce COD-VAE, a model that represents 3D shapes using a set of compact 1D latent vectors.
It replaces direct mappings of VecSet with a two-stage autoencoder scheme.
This allows our model to construct the compressed latent space while enabling efficient neural fields decoding.

The overall architecture of COD-VAE is illustrated in \Fref{fig:method_overview}.
Given an input point cloud $\mathcal{P}=\{\mathbf{p}_i\}_i^{N}$, our model encodes 1D latent vectors $\mathcal{Z} = \{\mathbf{z}_i\}_{i=1}^M$ from this point cloud.
In the decoding stage, the latent decoder first decompress $\mathcal{Z}$ along the channel dimension, which is then decoded to dense triplane patches.
Finally, we compute continuous occupancy values using the reconstructed triplanes and a shallow MLP.
The uncertainty-guided pruning further improves the efficiency of the decoder.
The following paragraphs describe details of each component.

\textblock{Encoder.}
Our encoder begins by obtaining initial positions of intermediate point patches and latent vectors.
Same as VecSet, these are sampled from the input point cloud using FPS.
The initial $C$-dimensional features of high-resolution points $\mathcal{G}^{(0)}=\{\mathbf{g}_i\}_{i=1}^N$, intermediate point patches $\mathcal{H}^{(0)}=\{\mathbf{h}_i \}_{i=1}^L$, and compact vectors $\mathcal{F}=\{\mathbf{f}_i\  \}_{i=1}^M$, are computed by applying positional embedding $\gamma(\cdot)$ to their positions.
Note that the sizes of the sets are $M \ll L \ll N$.

\begin{figure}[t]
    \centering
    \includegraphics[width=1\linewidth]{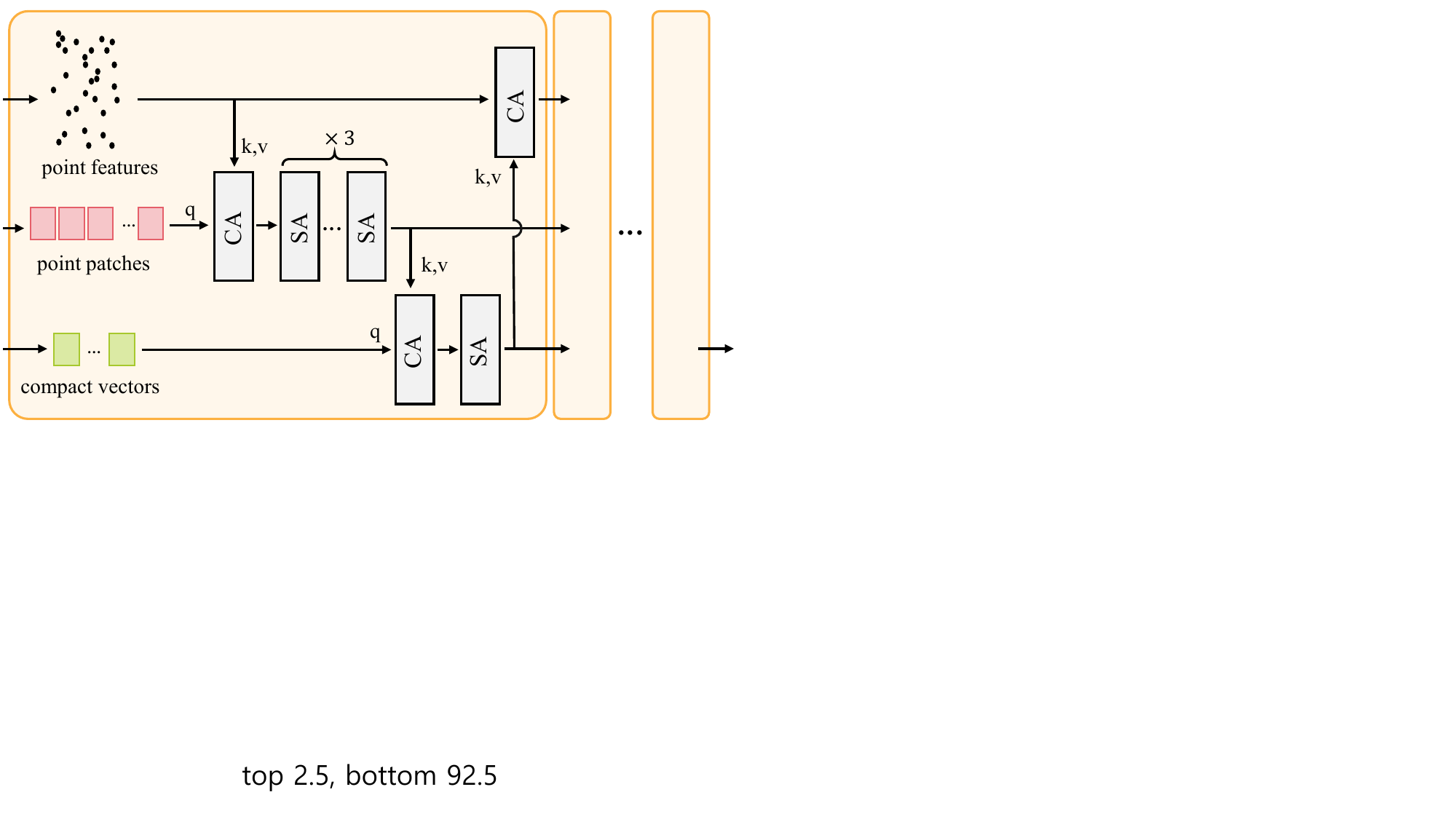}
    \vspace{-15pt}
    \caption{
    \textbf{Encoder block.}
    Our encoder block first projects high-resolution point features to the intermediate point patches and process them with self-attention layers. The processed patches are compressed into the compact 1D vectors.
    }
    \label{fig:method_encoder}
    \vspace{-10pt}
\end{figure}

\Fref{fig:method_encoder} illustrates the design of our encoder block.
Our encoder block progressive converts high-resolution point features to compact feature vectors through the intermediate point patches.
It first aggregates high-resolution features $\mathcal{G}$ to $\mathcal{H}$ through the cross-attention, and process the aggregated patches via self-attention layers. The processed patches are then mapped to $\mathcal{F}$.
Finally, the high-resolution features $\mathcal{G}^{(l)}$ are further enriched by projecting $\mathcal{F}$ onto them.
The outputs of $l$-th encoder block, $\mathcal{H}^{(l)}$, $\mathcal{F}^{(l)}$, $\mathcal{G}^{(l)}$, can be expressed as
\begin{equation}
    \mathcal{H}^{(l)} = \mathrm{SelfAttn}^3(\mathrm{CrossAttn}(\mathcal{H}^{(l-1)}, \mathcal{G}^{(l-1)})),
\end{equation}
\begin{equation}
    \mathcal{F}^{(l)} = \mathrm{SelfAttn}(\mathrm{CrossAttn}(\mathcal{F}^{(l-1)}, \mathcal{H}^{(l)})),
\end{equation}
\begin{equation}
    \mathcal{G}^{(l)} = \mathrm{CrossAttn}(\mathcal{G}^{(l-1)}, \mathcal{F}^{(l)}),
\end{equation}
where $\mathrm{SelfAttn}^{3}(\cdot)$ is a stack of 3 self-attention layers.

After the encoder blocks, the KL block transforms the compact feature vectors into 1D latent vectors.
Similar to \Eref{eq:vecset_kl}, it further compresses $\mathcal{F}$ along the channel dimension.
This process yields a set of compact latent vectors, $\mathcal{Z}=\{ \mathbf{z}_i \in \mathbb{R}^D \}_{i=1}^M$, where $D \ll C$.
These latent vectors are also regularized by KL divergence, same as VecSet.

\textblock{Latent decoder.}
The decoding process begins with the latent decoder, which decompresses $\mathcal{Z}$ along the channel dimension.
It consists of a linear layer that projects $\mathcal{Z}$ into $C$-dimensional features, followed by self-attention layers that produce the decompressed features $\mathcal{F}'$.
This process typically incurs negligible computation costs, as the size of our latent vector set is small.
The latent decoder is trained separately from the other components, with explicit training objectives focusing on channel decompression.
We provide further details of the two-stage training in \sref{sec:training}.

\textblock{Decoder.}
After the latent decoding, the decoder reconstructs dense triplane features from the decompressed vectors $\mathcal{F}'$ using transformer blocks.
While these intermediate triplanes enable effective and efficient neural fields decoding, they also increase computational costs of the decoder, scaling quadratic to the triplane resolutions.
To mitigate this, we introduce an uncertainty-guided token pruning that eliminates redundant computations in simple regions.

The architecture of our decoder is presented in \Fref{fig:method_overview}.
We use a learnable token sequence $\mathbf{e}\in \mathbb{R}^{(\frac{R}{f}\times\frac{R}{f}) \times C}$ to represent the positions of triplane tokens, where $R$ is the triplane resolution and $f$ is the patch size. 
We initialize the triplane tokens by querying from $\mathcal{F}'$ using a cross-attention block.
Then the auxiliary head predicts uncertainty values of triplane tokens.
We retain only the top $25\%$ tokens with the highest uncertainty values and prune the rest.
The remaining tokens are added with a shared mask token, concatenated with $\mathcal{F}'$, and processed by ViT-style transformer blocks.
Finally, we gather the processed triplane tokens and the pruned ones, which are projected by a linear layer to reconstruct full triplane features.

\textblock{Neural fields decoding.}
To reconstruct 3D objects, we decode continuous neural fields with the reconstructed triplane.
For a query point $\mathbf{q}$, we retrieve features from each plane and add these features to obtain a query-wise feature. This feature is passed through a shallow MLP and converted to an occupancy value.
Our triplane-based neural fields decoding process achieves substantial improvement in efficiency compared to VecSet, as it only requires computations for bilinear interpolation and a shallow MLP.

\begin{figure}[t]
    \centering
    \includegraphics[width=1\linewidth]{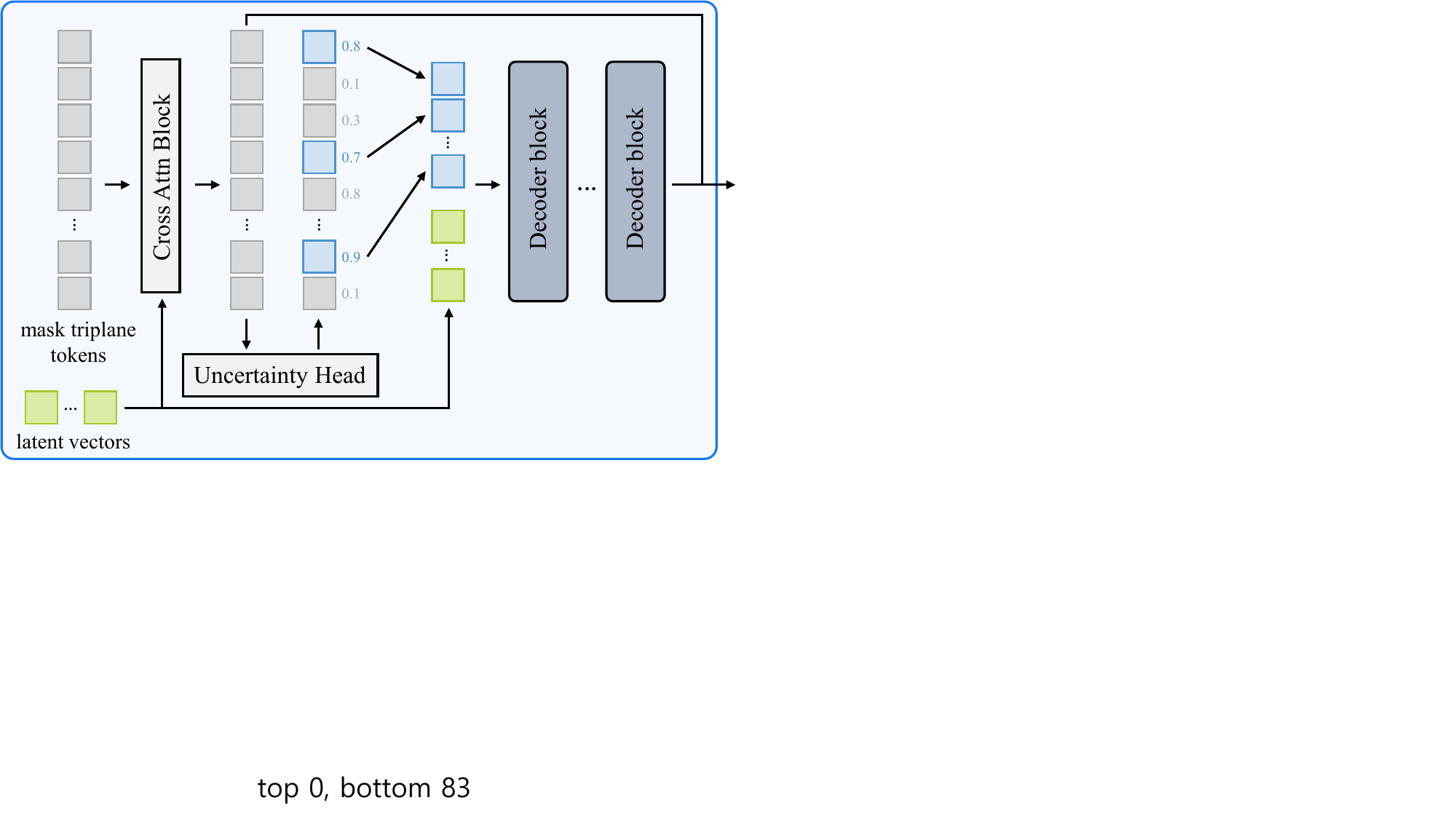}
    \vspace{-15pt}
    \caption{
    \textbf{Decoder design.}
    Our decoder prunes most of the triplane tokens at the beginning based on the predicted uncertainty values. The processed and pruned tokens are gathered at the last stage of the decoder to reconstruct the full triplane features.
    }
    \label{fig:method_decoder}
    \vspace{-10pt}
\end{figure}

\subsection{VAE training}
We train our COD-VAE using binary cross entropy loss, with KL divergence loss to regularize latent vectors and uncertainty loss to optimize the uncertainty head.
Please refer to the supplements for the full objective of our model.

\label{sec:training}
\textblock{Uncertainty head.}
The uncertainty head is trained to predict reconstruction error, which is computed using the initial triplane tokens.
We first apply a sigmoid activation to the predicted uncertainty values and arrange them into triplanes.
Then, we compute the binary cross entropy loss $\mathcal{L}_{rec}(\mathbf{q})$ for a query point $\mathbf{q}$.
Following \cite{fridovich2023kplanes}, we retrieve uncertainty values from the triplanes and compute the query-wise uncertainty $u(\mathbf{q})$ as
\begin{equation}
    u(\mathbf{q}) = \psi_{xy}(U_{xy}, \mathbf{q}) \cdot \psi_{yz}(U_{yz}, \mathbf{q}) \cdot \psi_{xz}(U_{xz}, \mathbf{q}),
\end{equation}
where $\psi_{C}$ projects $\mathbf{q}$ onto $C$-plane and retrieves the corresponding feature, and $U_C\in\mathbb{R}^{\frac{R}{f} \times \frac{R}{f}}$ is the triplane uncertainty on the $C$-plane.
The uncertainty head is trained by minimizing the mean-squared-error (MSE) between $u(\mathbf{q})$ and $\mathcal{L}_{rec}(\mathbf{q})$.
This objective encourages the uncertainty head to identify tokens that have higher reconstruction errors and need more computations.
Our uncertainty-guided pruning eliminates redundant computations in simple regions, thereby addressing computation costs of the decoder.

\textblock{Two-stage training.}
Instead of jointly training all components, we adopt a two-stage training strategy.
In the first stage, we train the autoencoder, a model without the KL block and the latent decoder.
In the second stage, we freeze the trained components and optimize the KL block and the latent decoder by minimizing the MSE loss between the outputs of the latent decoder, $\mathcal{F}'$, and the encoder outputs, $\mathcal{F}$.
We also minimize the KL divergence loss, as well as the reconstruction error for stable training.
By encouraging the latent decoder to specialize in channel decompression, the two-stage training strategy improves the accuracy of VAEs.

\subsection{Diffusion model}
We train a latent diffusion model on the compact latent space constructed by our COD-VAE.
Following VecSet \cite{zhang2023vecset}, we adopt transformer-based denoising networks and use the diffusion formulations from EDM \cite{karras2022edm}.
While we do not explicitly aim to improve the generation process itself, the compact latent space provided by COD-VAE significantly enhances the efficiency of the diffusion models.
\section{Experiments}
To verify the effectiveness of our method, we first evaluate the reconstruction quality of COD-VAE.
We then assess the performance of diffusion models on both class-conditioned and unconditional generation tasks.
Additional details of the experiments and results are provided in the supplements.

\textblock{Datasets.}
We conduct the experiments on two public datasets, ShapeNet-v2 \cite{jun2023shap} and Objaverse \cite{deitke2023objaverse}.
ShapeNet \cite{chang2015shapenet} consists of 3D CAD models from 55 categories.
We follow the experimental setups of \cite{zhang20223dilg, zhang2023vecset}, and use same train/validation splits and preprocessed meshes.

Objaverse contains around 80K objects, which are more diverse and complex than the objects in ShapeNet.
Due to its large size, we sample the subset of around 150K objects based on the list provided in \cite{qiu2024richdreamer}.
We split the dataset into 135,511 objects for training and 14,663 for validation.
Following \cite{zhang2024lagem}, we employ ManifoldPlus \cite{huang2020manifoldplus} to make watertight meshes, and normalize them to be in unit cubes.
We use the full dataset for training VAE, and use the subset with around 90K objects for training diffusion models.

\textblock{Baselines.}
We select VecSet \cite{zhang2023vecset} as our primary baseline and report its results for various numbers of latent vectors $M$.
We also compare with additional state-of-the-art methods, including 3DILG \cite{zhang20223dilg}, GEM3D \cite{petrov2024gem3d}.
We reproduce results of VecSet using its official code, with the latent channel dimension $D=32$.
We report the results of other baselines using either the provided pretrained weights, or their official code when the weights are unavailable.

\textblock{Evaluation metrics.}
For the reconstruction experiments, we follow VecSet and use the mean intersection-over-union (IoU) of 50K query points uniformly sampled from unit cubes.
We also report Chamfer Distance (CD) and F-score (F1), computed from 10K points sampled from both the reconstructed and ground truth meshes.
For the generation experiments, we adopt FID-based scores as our main metrics due to their reliability.
Following \cite{zheng2022sdf, zhang2023vecset, petrov2024gem3d}, we measure the FID of the images rendered from 20 viewpoints (Rendering-FID), as well as the Fréchet PointNet++ Distance using features from PointNet++ (Surface-FPD).
To compute FPD, we train PointNet++ \cite{qi2017pointnet++} on ShapeNet-55 classification and use this model to extract features.
We additionally measure MMD, COV, and 1-NNA, following \cite{zheng2022sdf}.
MMD assess the fidelity, COV measures the coverage, and 1-NNA evaluates the diversity of the generated shapes.
We measure the throughput using a single A6000 GPU.

\textblock{Implementation details.}
Our model consists of 4 encoder blocks and 12 decoder layers, with $C=512$ and $D=32$.
We set the patch size $f=8$ for the decoder in all experiments.
During training, we sample 4,096 query points uniformly from unit cubes and 4,096 points near the objects' surfaces.
We train the autoencoders for 1,600 epochs on ShapeNet and 300 epochs on Objaverse.
The second-stage VAE training follows the same number of epochs.
Our models are trained using 4 RTX 4090 GPUs for ShapeNet and 4 A6000 GPUs for Objaverse.
Training VAEs takes about 3 days for ShapeNet and 6 days for Objaverse, and training diffusion models takes less than 4 days for both datasets.

\begin{figure*}[t]
    \centering
    \includegraphics[width=1\linewidth]{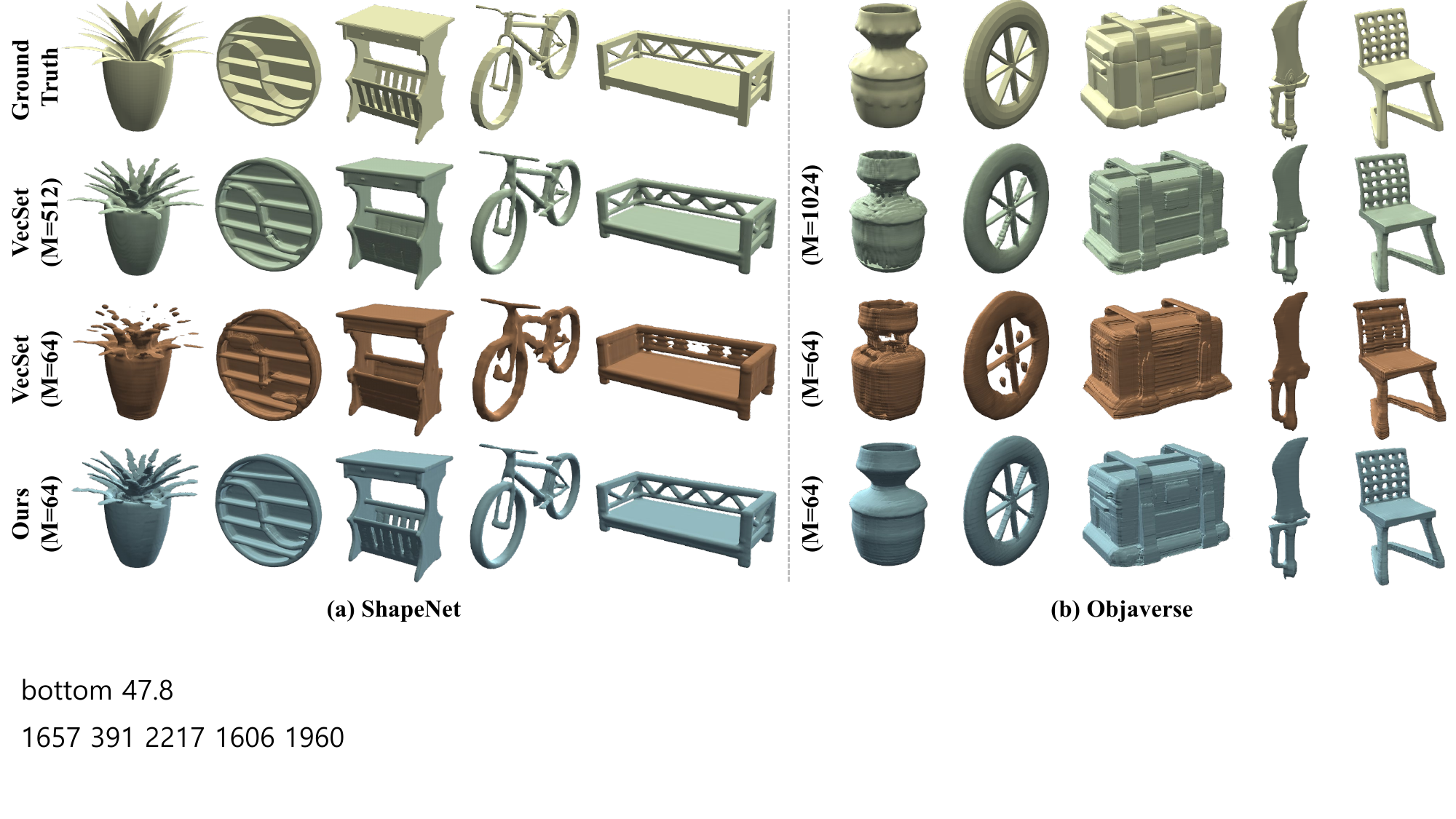}
    \vspace{-15pt}
    \caption{\textbf{Qualitative reconstruction results.} Comparisons of VAEs on ShapeNet (a) and Objaverse (b) are presented.}
    \label{fig:qualitative_recon}
    \vspace{-10pt}
\end{figure*}
\subsection{Reconstruction results}
For ShapeNet experiments, we follow previous research \cite{zhang20223dilg, zhang2023vecset} and report results of both autoencoders (without channel compression by KL block) and VAEs.
We also set the input point cloud size $N=2048$, the test grid resolution to 128.
For our models, we set $L=512$, and the triplane resolution to 128.
For Objaverse, due to the complexity of the objects, we increase the input point cloud size to $N=8192$.
We increase the number of latent vectors of VecSet by 2 as well as the number of our point patches ($M=L=1024$), and the triplane resolution to 256.

\begin{table}[]
\setlength{\tabcolsep}{2pt}
\centering
\footnotesize
\begin{tabular}{c|ccc|ccc}
\toprule
 & \multicolumn{3}{c}{Autoencoder} & \multicolumn{3}{c}{VAE} \\
 Method & IoU (\%)$\uparrow$ & CD$\downarrow$ & F1 (\%)$\uparrow$ & IoU (\%)$\downarrow$ & CD$\downarrow$ & F1 (\%)$\uparrow$  \\
\midrule
3DILG (M=512) & 95.9 & 0.013 & 98.0 & 94.7 & 0.013 & 97.6 \\
VecSet (M=32) & 87.8 & 0.021 & 91.3 & 87.0 & 0.021 & 90.6 \\
VecSet (M=64) & 91.2 & 0.017 & 94.7 & 90.9 & 0.017 & 94.4 \\
VecSet (M=512) & 96.3 & 0.013 & 98.0 & 96.2 & 0.013 & \textbf{98.0} \\
Ours (M=32) & 96.1 & \textbf{0.012} & 98.0 & 95.8 & 0.013 & 97.8 \\
Ours (M=64) & \textbf{96.5} & \textbf{0.012} & \textbf{98.2} & \textbf{96.3} & \textbf{0.012} & \textbf{98.0} \\
\bottomrule

\end{tabular}
\vspace{-5pt}
\caption{
\textbf{Reconstruction results on ShapeNet.}
}
\label{table:shapenet_recon}
\vspace{-10pt}
\end{table}
\textblock{Results on ShapeNet.}
\Tref{table:shapenet_recon} reports the reconstruction results on ShapeNet.
Our models, both autoencoders and VAEs, outperform all other methods with only using 64 latent vectors. Even with 32 latent vectors, our models achieve reconstruction quality comparable to VecSet ($M=512$), offering a more efficient yet effective option for diffusion models.
In contrast, simply reducing the number of latent vectors of VecSet leads to noticeable degradation in both autoencoders and VAEs.
This is also evident in the qualitative results presented in \Fref{fig:qualitative_recon} (a).
Thanks to the two-stage autoencoding, our models with $M=64$ achieve high-quality reconstruction results, whereas VecSet with $M=64$ fails to represent fine details of the objects.

\textblock{Results on challenging objects.}
We then evaluate the performance on Objaverse, which comprises more challenging and diverse objects than ShapeNet.
As reported in \Tref{table:objaverse}, our model with $M=64$ exhibits even higher reconstruction quality than VecSet with $M=1024$.
Our VAE also significantly boosts the efficiency of the generation, achieving $21.6\times$ higher throughput than VecSet with $M=1024$.
We also deliver the qualitative results in \Fref{fig:qualitative_recon} (b).
Reducing the number of latent vectors of VecSet leads to the severe degradation.
These results showcase that we can represent 3D shapes in high-quality using 64 latent vectors.

\subsection{Generation results}
\label{sec:generation}
To further validate our latent representations, we evaluate the generation performance of diffusion models trained on our compact latent space.
We deliver the results of class-conditioned and unconditional generation, which are conducted on ShapeNet and Objaverse, respectively.
We follow VecSet and set the number of sampling steps 18 for ShapeNet.
For Objaverse, we increase the sampling steps to 50 as it comprises more diverse and complex objects.

\begin{table}[]
\setlength{\tabcolsep}{2pt}
\centering
\footnotesize
\begin{tabular}{c|ccc|cc}
\toprule
 & \multicolumn{3}{c}{VAE} & \multicolumn{2}{c}{Generation} \\
 Method & IoU (\%)$\uparrow$ & CD$\downarrow$ & F1 (\%)$\uparrow$ & Throughput & Mem.  \\
\hline
VecSet ($M=64$) & 71.2 & 0.054 & 85.2 & 1.99 & 5.46 \\
VecSet ($M=1024$) & 79.8 & 0.051 & 87.0 & 0.23 & 9.96 \\
Ours ($M=64$) & \textbf{79.9} & \textbf{0.046} & \textbf{88.0} & \textbf{4.97} & \textbf{3.08} \\
\bottomrule

\end{tabular}
\vspace{-5pt}
\caption{
\textbf{Reconstruction results on Objaverse.} The GPU memory (GB) is measured with a batch size 16.
}
\label{table:objaverse}
\vspace{-10pt}
\end{table}



\begin{table*}[]
\setlength{\tabcolsep}{6pt}
\centering
\footnotesize
\begin{tabular}{c|cc|cc|cc|cc|cc}
\toprule
\multirow{2}{*}{Method} & \multirow{2}{*}{Rendering-FID $\downarrow$} & \multirow{2}{*}{Surface-FPD $\downarrow$} & \multicolumn{2}{c}{MMD $\downarrow$} & \multicolumn{2}{c}{COV (\%) $\uparrow$} & \multicolumn{2}{c}{1-NNA (\%) $\downarrow$} & \multicolumn{2}{c}{Throughput (sample/s) $\uparrow$} \\
    & &  & CD & EMD & CD & EMD & CD & EMD & Sampling & Full  \\
\hline

3DILG \cite{zhang20223dilg} & 64.83 & 2.585 & 6.270 & 10.98 & 58.16 & 62.80 & 61.49 & 59.90 & 0.14 & 0.09 \\
GEM3D \cite{petrov2024gem3d} & 47.98 & 0.997 & 5.381 & 10.41 & 66.64 & 66.05 & 54.74 & 56.02 & 0.09 & 0.07 \\ 
VecSet ($M=32$) \cite{zhang2023vecset} & 65.56 & 0.800 & 4.999 & 10.44 & 84.51 & 82.85 & 54.32 & 54.76 & 39.77 & 2.81 \\
VecSet ($M=64$) & 54.47 & 0.629 & 5.090 & 10.49 & 84.89 & 84.45 & 55.04 & 54.94 & 21.92 & 2.63 \\
VecSet ($M=512$) & 44.18 & 0.521 & \textbf{4.807} & \textbf{10.21} & 85.20 & 84.23 & 53.78 & 55.06 & 2.59 & 1.16 \\ 
Ours ($M=32$) & 37.34 & 0.473 & 5.020 & 10.38 & 84.96 & 84.23 & \textbf{53.11} & 54.09 & \textbf{41.19} & \textbf{24.17}\\
Ours ($M=64$) & \textbf{37.05} & \textbf{0.460} & 5.025 & 10.38 & \textbf{86.03} & \textbf{85.37} & 54.89 & \textbf{54.07} & 22.49 & 16.09 \\
\bottomrule

\end{tabular}
\vspace{-5pt}
\caption{
\textbf{Class-conditioned generation results on ShapeNet.}
The scales of MMD are $10^{-3}$, and $10^{-2}$ for CD, and EMD, respectively.
The `Sampling' denotes the throughput of the latent sampling, and the `Full' denotes the throughput of the entire process.
}
\label{table:shapenet_gen}
\vspace{-10pt}
\end{table*}


\begin{table}[]
\setlength{\tabcolsep}{6pt}
\centering
\footnotesize
\begin{tabular}{c|cc|cc}
\toprule
 & \multicolumn{2}{c}{Batch size 32} & \multicolumn{2}{c}{Batch size 64} \\
 & Speed & Memory & Speed & Memory \\
\hline
VecSet ($M=512$) & 1.01 & 39.01 & - & OOM \\
Ours ($M=32$) & 2.60 & 5.17 & 1.37 & 8.33 \\
Ours ($M=64$) & 2.23 & 5.63 & 1.20 & 9.23 \\
\bottomrule

\end{tabular}
\vspace{-5pt}
\caption{
\textbf{Training efficiency of the diffusion models.} We report the speed (iter/s) and the GPU memory (GB) on ShapeNet.
}
\label{table:efficiency}
\vspace{-10pt}


\end{table}
\begin{figure*}[t]
    \centering
    \includegraphics[width=1\linewidth]{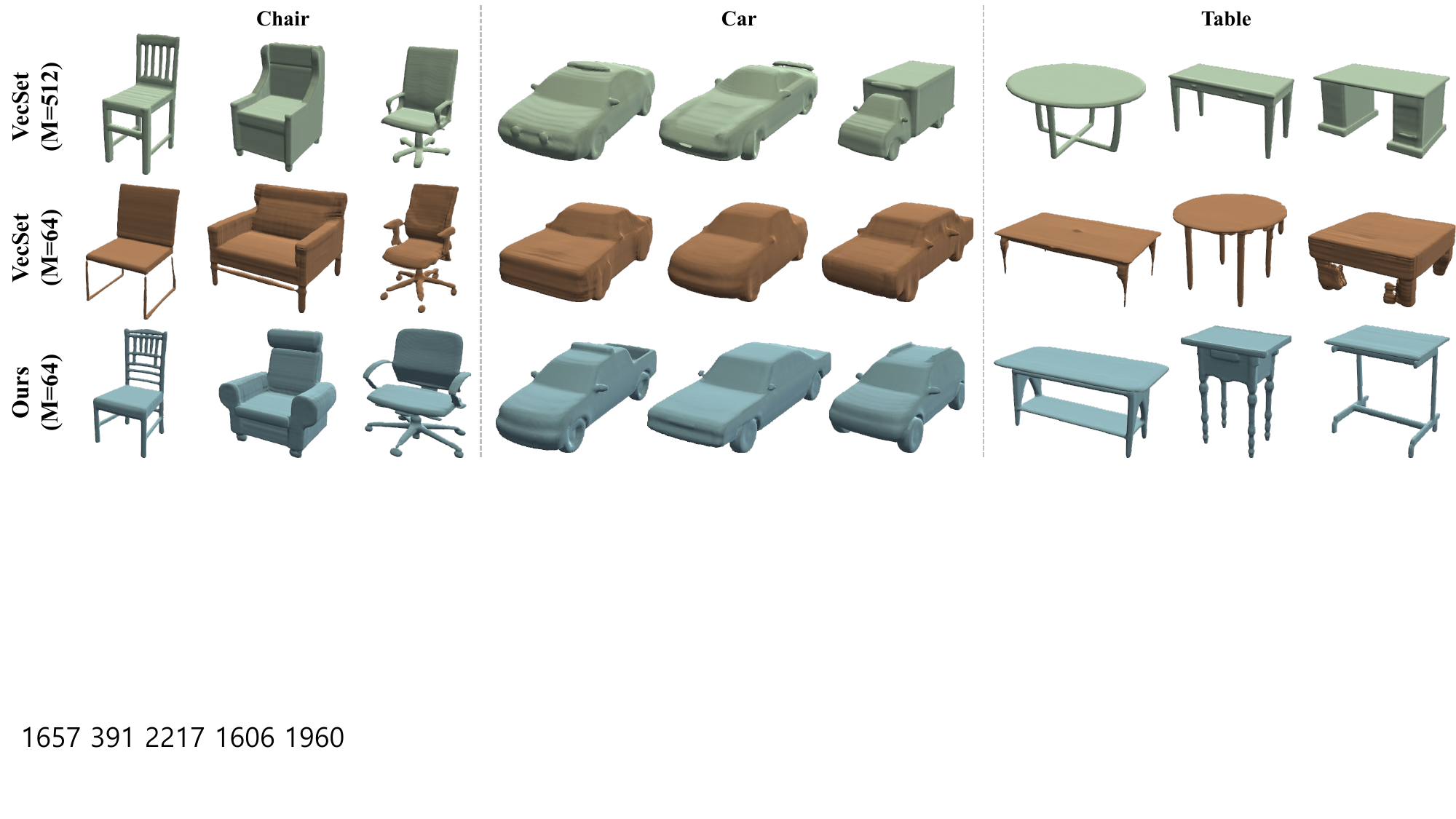}
    \vspace{-15pt}
    \caption{\textbf{Qualitative class-conditioned generation results on ShapeNet.}}
    \label{fig:qualitative_class_cond}
    \vspace{-10pt}
\end{figure*}

\textblock{Category-conditioned generation.}
\Tref{table:shapenet_gen} reports the class-conditioned generation results.
The results are averaged over 5 commonly used classes, which are \textit{airplane}, \textit{car}, \textit{chair}, \textit{table}, and \textit{rifle}.
Our models achieve the highest performance in FID-based scores (Rendering-FID, Surface-FPD) and outperform other methods in most additional metrics.
In contrast, VecSet with $M=32,64$ delivers degraded results, as reflected in FID-based scores and further shown in \Fref{fig:qualitative_class_cond}.
Notably, our method significantly improves the efficiency of generative models, achieving $13.9\times$ and $20.84\times$ improved throughput.
The compact latent space also improves the training efficiency, as reported in \Tref{table:efficiency}.


\begin{figure*}[t]
    \centering
    \includegraphics[width=1\linewidth]{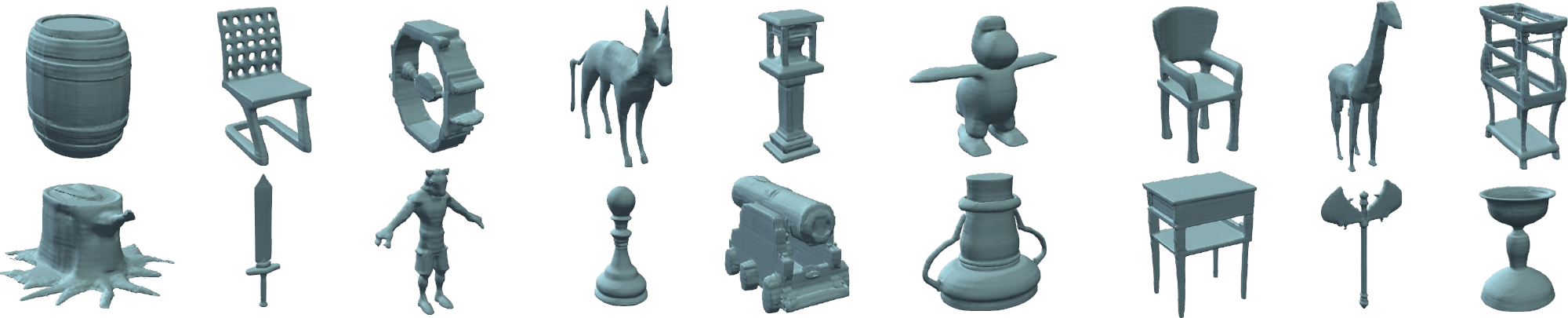}
    \vspace{-17pt}
    \caption{\textbf{Unconditional generation results.} Results generated by our model with $M=64$ are presented.}
    \label{fig:qualitative_uncond}
    \vspace{-5pt}
\end{figure*}
\textblock{Unconditional generation.}
We present that our diffusion model, trained on the compact latent space, can learn the distributions of a wide range of objects.
We deliver the qualitative unconditional generation results of our model in \Fref{fig:qualitative_uncond}.
Our model generates diverse types of shapes in high-quality with significantly improved efficiency.

\subsection{Ablation study}
\label{sec:ablation}
To examine the effects of each proposed component, we conduct ablation studies on ShapeNet reconstruction.
All ablations are performed using autoencoders with $M=32$.

\textblock{Encoder design.}
We first replace our encoder with the encoder of VecSet, which utilizes the cross-attention to directly map point features to latent vectors and self-attention layers to process latent vectors.
We compare against models with encoders having various number of cross-attention layers.
As shown in \Tref{table:ablation} (a), our encoder outperforms the models with VecSet's encoders by a large margin.
Using more cross-attention layers does not improve performance beyond a certain point.
This highlights the importance of intermediate point patches, which enables higher compression ratio by progressive transformations.

\textblock{Decoder design.}
We also replace our decoder with the cross-attention layer as used in VecSet, which directly maps latent vectors to query points.
Ablation results on the decoder design is presented in \Tref{table:ablation} (b).
Our decoder improves the overall reconstruction quality as well as efficiency, compared to that of VecSet.
In addition, the intermediate triplane representations significantly reduces computation costs of the neural fields decoding, which accompanies more than 20M query points to reconstruct $128^3$ grids.
Adding more cross-attention layers could offer better reconstruction quality, but at the cost of excessive computations.

\begin{table*}[h]
    \centering
    \footnotesize
    \begin{subtable}[t]{0.31\textwidth}
        \vspace{0pt}
        \centering
        \setlength{\tabcolsep}{3pt}
        \begin{tabular}{c|ccc}
        \toprule
         Encoder & IoU (\%)$\uparrow$ & CD$\downarrow$ & F1 (\%)$\uparrow$  \\
        \hline
        CrossAttn & 91.2 & 0.016 & 94.7 \\
        CrossAttn $\times12$ & 93.4 & 0.014 & 96.5 \\
        CrossAttn $\times24$ & 93.5 & 0.014 & 96.6 \\
        Ours & 96.1 & 0.012 & 98.0 \\
        \bottomrule
        \end{tabular}
        \vspace{0.5pt}
        \caption{Encoder design.}
    \end{subtable}
    \hfill
    \begin{subtable}[t]{0.4\textwidth}
        \vspace{0pt}
        \centering
        \setlength{\tabcolsep}{3pt}
        \begin{tabular}{c|cccc}
        \toprule
         \multirow{2}{*}{Decoder} & \multirow{2}{*}{IoU (\%)$\uparrow$} & \multirow{2}{*}{CD$\downarrow$} & \multirow{2}{*}{F1 (\%)$\uparrow$} & Recon.  \\
          & & & & (sample/s)$\downarrow$ \\
        \hline
        CrossAttn & 95.8 & 0.013 & 97.8 & 3.43 \\
        CrossAttn$\times3$ & 96.3 & 0.013 & 98.0 & 1.47 \\
        Triplane & 96.1 & 0.012 & 98.0 & 42.68 \\
        \bottomrule
        \end{tabular}
    \caption{Decoder design.}
    \label{tab:ablation_decoder}
    \end{subtable}
    \hfill
    \begin{subtable}[t]{0.28\textwidth}
        \vspace{0pt}
        \centering
        \setlength{\tabcolsep}{3pt}
        \begin{tabular}{c|cccc}
        \toprule
         Method & IoU (\%)$\uparrow$ & CD$\downarrow$ & F1 (\%)$\uparrow$  \\
        \hline
        Prune 0\% & 96.2 & 0.012 & 98.0 \\
        Prune 50\% & 96.2 & 0.012 & 98.0 \\
        Prune 75\% & 96.1 & 0.012 & 98.0 \\
        Prune 90\% & 95.9 & 0.012 & 98.0 \\
        \bottomrule
        \end{tabular}
        \vfill
        \caption{Pruning ratio results with $R=128$.}
    \end{subtable}
    \vspace{-8pt}
    \caption{\textbf{Quantitative ablation results} with $M=32$ on ShapeNet reconstruction.}
    \vspace{-10pt}
    \label{table:ablation}
\end{table*}

\begin{table}[]
\setlength{\tabcolsep}{7pt}
\centering
\footnotesize
\begin{tabular}{c|cc|cc}
\toprule
Pruning & \multicolumn{2}{c}{Speed (iter/s)} & \multicolumn{2}{c}{Memory (GB)} \\
 ratio & $R=128$ & $R=256$ & $R=128$ & $R=256$ \\
\hline
0\% & 4.21 & 1.17 & 6.79 & 26.99 \\
50\% & 5.14 & 2.21 & 5.50 & 11.87 \\
75\% & 5.60 & 3.09 & 5.02 & 7.80 \\
90\% & 5.78 & 3.71 & 4.78 & 6.33 \\
\bottomrule

\end{tabular}
\vspace{-5pt}
\caption{
\textbf{Training efficiency analysis of the pruning.}
}
\label{table:ablation_pruning}
\vspace{-5pt}
\end{table}

\textblock{Pruning ratio.}
Finally, we validate our uncertainty-guided pruning by changing the pruning ratio.
As shown in \Tref{table:ablation} (c) and \Tref{table:ablation_pruning}, training costs are effectively reduced as the pruning ratio increases with slight performance drops.
With $R=128$,  pruning 50\% triplane tokens maintains the reconstruction quality while achieving $1.22\times$ speedup and $0.81\times$ memory consumption for training the autoencoders.
Further increasing the pruning ratio to 75\% results in a $1.33\times$ speedup and $0.73\times$ memory, with only a slight drop in performance, which we choose as our main setting.
The benefits of our uncertainty-guided pruning become more evident as the triplane resolution increases to $R=256$.
This improved efficiency allows our model to scale more effectively to datasets with more complex samples, \eg, Objaverse.
\section{Discussion}

\textblock{Accelerating the query decoding.}
Reconstructing neural fields typically requires evaluating a vast number of query point (over 2M points for $128^3$ test grids).
While our triplane-based query decoding significantly reduces these overheads, it often hinders maximizing throughput (see \Tref{table:shapenet_gen}, Ours with M=32).
This can be further addressed by utilizing the improved query points sampling, \eg, multi-resolution approach as proposed in \cite{mescheder2019occnet}.
We further provide runtime analysis with this technique in the supplements.

\begin{figure}[t]
    \centering
    \includegraphics[width=1\linewidth]{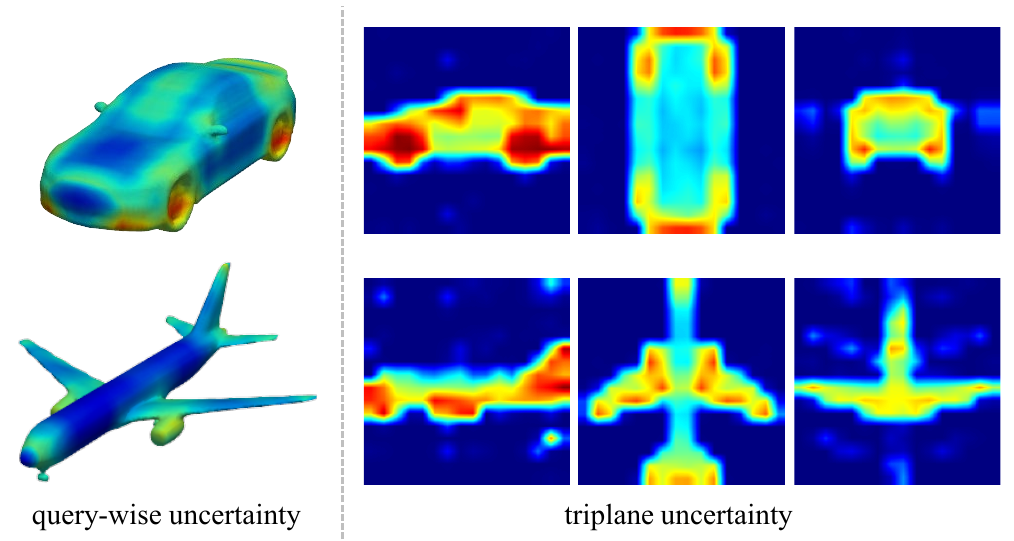}
    \vspace{-15pt}
    \caption{\textbf{Visualizations of the predicted uncertainty values.}}
    \label{fig:uncertainty_vis}
    \vspace{-10pt}
\end{figure}

\textblock{Uncertainty visualization.}
The predicted uncertainty values serve as the importance score of the triplane tokens.
As presented in \Fref{fig:uncertainty_vis}, our uncertainty head assigns higher scores to more detailed regions, placing more computations on them.
This allows our model to avoid heavy computations for simple parts, and place more computational resources on reconstructing details and thin structures.

\begin{table}[]
\setlength{\tabcolsep}{3pt}
\centering
\footnotesize
\begin{tabular}{c|cccccc}
\toprule
Num. latents (M) & 64 & 128 & 256 & 512 & 1024 & 2048 \\
\hline
Memory (GB) & 1.13 & 1.64 & 2.86 & 5.65 & 13.63 & 38.66 \\
Max batch size & 128 & 64 & 32 & 16 & 4 & 2 \\
Throughput (sample/s) & 652.45 & 319.00 & 149.74 & 65.16 & 24.95 & 8.37 \\
\bottomrule

\end{tabular}
\vspace{-5pt}
\caption{
\textbf{Running statistics of the denoising network} with various $M$.
The network has 24 layers with 512 channel.
The GPU memory is measured with a batch size 2, and the throughput is measured with the corresponding maximum batch size.
}
\label{table:runtime_scaling}
\vspace{-10pt}
\end{table}
\textblock{Scaling and the compact latent space.}
The efficiency improvement of the compact latent space becomes more apparent as the number of latent vectors increases, as shown in \Tref{table:runtime_scaling}.
With the same compression ratio $8\times$, reducing $M$ from 512 to 64 leads to the $10.0\times$ throughput and $0.20\times$ memory, while reducing from 2048 to 256 results in the $17.9\times$ throughput and $0.07\times$ memory usage.
We highlight that constructing the compact latent space can be a key for scaling generative models to more complex scenes, which usually requires a large number of latent vectors. 

\textblock{Limitations and future works.}
While our COD-VAE successfully alleviates bottlenecks of the generation process, parameter counts of VAE and computations of the encoder are increased compared to VecSet (see the supplements for more details).
We anticipate that enhancing the encoder efficiency can further boost up the training of VAEs and generative models.
Moreover, our VAE has the potentials to encode information from multiple sources, \eg, images and videos.
This can be an interesting avenue for future works.
\section{Conclusion}
Reducing the latent size is the key for efficient 3D generation.
Our COD-VAE with a two-stage autoencoding scheme represents 3D shapes with 64 latent vectors without sacrificing reconstruction and generation quality.
This compact latent vectors set enables efficient and high-quality generation, suggesting a new direction for 3D generative models.

\clearpage
\section*{Acknowledgments}
This work was partly supported by Institute of Information \& communications Technology Planning \& Evaluation (IITP) grant funded by the Korea government(MSIT) (No. RS-2024-00457882), and Artificial Intelligence Graduate School Program, Yonsei University, under Grant RS-2020-11201361.

{
    \small
    \bibliographystyle{ieeenat_fullname}
    \bibliography{main}
}
\clearpage
\clearpage
\maketitlesupplementary
\appendix

\section{Additional discussion}

\begin{table}[]
\setlength{\tabcolsep}{4pt}
\small
\begin{tabular}{c|ccc}
\toprule
 & Sampling & Decoding & Full \\
\hline
VecSet ($M=512$) & 2.40 & 2.10 & 1.12  \\
Ours ($M=64$) & 20.49 & 56.44 & 15.03 \\
Ours ($M=32$) & 37.42 & 58.22 & 22.78 \\
\hline
VecSet ($M=512$) with MR & 2.40 & 10.61 & 1.94  \\
Ours ($M=64$) with MR & 20.67 & 162.85 & 18.34 \\
Ours ($M=32$) with MR & 37.70 & 174.88 & 31.00 \\
\bottomrule

\end{tabular}
\vspace{-5pt}
\caption{
\textbf{Generation throughput (sample/s) at each step.} `Sampling' denotes the latent generation, `Decoding' denotes the latent-to-fields decoding process, including both the decoder and the neural fields decoding, and `Full' denotes the entire generation process.
`MR' denotes the multi-resolution query points sampling technique.
The throughput is measured using a single RTX A6000 GPU with a batch size 64.
}
\label{table:supp_query_runtime}
\vspace{-10pt}
\end{table}
\subsection{Improved query points sampling}
As discussed in the main paper, we follow VecSet \cite{zhang2023vecset} and adopt a naive query point sampling approach, which employs a single-resolution test grid and evaluates all points within it. This results in an enormous number of query point evaluations.
While our triplane-based neural field decoding significantly reduces computational overhead in the query decoding process, the excessive number of query points often hinders maximizing the generation throughput.

The overall throughput can be further accelerated by adopting an improved query points sampling technique.
We can employ a simple multi-resolution query point sampling strategy, similar to the method proposed in \cite{mescheder2019occnet}.
For instance, in the case of a $128^3$ test grid, we first use coarse-level points from a $64^3$ grid to identify occupied and empty regions. Based on this coarse-level evaluation, we then evaluate finer-level points ($128^3$) only within occupied voxels.

As shown in \Tref{table:supp_query_runtime}, this strategy reduces unnecessary computations, leading to further acceleration in the neural field decoding process.
We can boost up our model with $M=32$ by alleviating the bottleneck in the neural fields decoding process, thereby maximizing its throughput.
While the same sampling technique can also be applied to VecSet ($M=512$), its computationally intensive sampling procedures impose a substantial overhead, limiting the efficiency gains in the generation process.
Notably, this method can be implemented as batch-wise computations by using an appropriate zero-padding.

\subsection{Reducing latent vectors and generation}
As shown in Tab.~3 in the main paper, compact latent vectors not only improve efficiency but also lead to better generation performance in all metrics except MMD.
We attribute this to the fact that a smaller total latent size simplifies the training of diffusion models, thereby enhancing overall generation quality.
Specifically, employing more latent vectors allows each vector to cover a smaller region, resulting in higher fidelity (MMD) of VecSet ($M=512$).
However, this comes at a cost of a larger total latent size, making diffusion models more challenging to model and resulting in degraded performance on all other metrics.
This results in degraded performance of VecSet with $M=512$ across all other metrics.
While more latent vectors may offer better higher fidelity, they also impose a greater burden on the diffusion model to capture the full distribution.
Similarly, recent diffusion models also employ VAEs for latents with a small channel dimension to make the total latent size small. 

\subsection{Future extension of COD-VAE}
\paragraph{Integration with VecSet-based methods.}
Recently, many feed-forward 3D diffusion models have extended the VecSet pipeline with various improvements.
GaussianAnything \cite{yushi2025ga} introduces a Gaussian-based decoder and adopts FPS-based 1D latent vectors, similar to those used in VecSet.
Dora \cite{chen2024dora} addresses the issue of uniform point sampling in VecSet by proposing a sharp-edge sampling strategy, which better captures regions with complex geometry.
Since COD-VAE also builds upon the VecSet framework, we believe our model can naturally benefit from these recent advances.

\paragraph{Towards textured mesh reconstruction.}
While this work follows the experimental setups of VecSet and only reconstructs shapes of the objects, extending our model to generate textured meshes can be an interesting future direction.
This can possibly be accomplished by employing multi-view diffusion models to generate materials and textures \cite{zhang2024clay}, training VAE with an additional rendering loss \cite{lan2024ln3diff, yushi2025ga}, or employing Gaussian-based VAE decoder \cite{yushi2025ga}.

\begin{table}[]
\setlength{\tabcolsep}{5pt}
\footnotesize
\begin{tabular}{c|ccc}
\toprule
 & Speed (iter/s)$\uparrow$ & Memory (GB)$\downarrow$ & Params.$\downarrow$ \\
\hline
VecSet ( $M=512$) & 2.98 & 13.40 & 113M  \\
Ours ($M=64$) & 3.02 & 15.64 & 187M \\
Ours ($M=32$) & 3.18 & 14.81 & 187M \\
\bottomrule

\end{tabular}
\vspace{-5pt}
\caption{
\textbf{Training efficiency analysis.} We analyze the efficiency of VAEs on ShapeNet, without the two-stage training strategy for a clear comparison.
}
\label{table:supp_vae_training}
\vspace{-10pt}
\end{table}
\begin{table}[]
\setlength{\tabcolsep}{6pt}
\footnotesize
\begin{tabular}{c|ccc}
\toprule
 & Encoder & Decoder & Fields decoding \\
\hline
VecSet ($M=512$) & 0.02 & 0.27 & 13.60  \\
Ours ($M=64$) & 0.30 & 0.10 & 0.46 \\
Ours ($M=32$) & 0.30 & 0.08 & 0.46 \\
\bottomrule

\end{tabular}
\vspace{-5pt}
\caption{
\textbf{Runtime breakdown of VAEs.} We report the latency (s) with a batch size 64.
}
\label{table:supp_vae_breakdown}
\vspace{-10pt}
\end{table}
\section{VAE efficiency analysis}
We present the training efficiency analysis of VAEs in \Tref{table:supp_vae_training}.
While our model with $M=64$ achieves a comparable training speed to VecSet, it requires slightly more GPU memory ($1.16\times$).
Additionally, the total parameter count increases by a factor of 1.65.
These increases in memory usage and parameter count can be viewed as the trade-offs for achieving improved generative efficiency.
Although VAEs are less frequently trained compared to generative models in practice, we believe that reducing these costs remains an interesting direction for future research.

We also provide a runtime breakdown of the inference of VAEs  in \Tref{table:supp_vae_breakdown}. Our models demonstrate greater efficiency in the decoder and the neural fields decoding. However, the computational cost of the encoder is higher compared to VecSet, as VecSet employs a lightweight encoder consisting of a single cross-attention layer.
Considering that the forward computation of the encoder is usually included in the generative model training, reducing the computations of the encoder can further accelerate the training of generative models, which we leave for future works.

\section{Method details}
In this section, we provide further details of our COD-VAE architecture design.
We provide details of each component in the following paragraphs.

\paragraph{Encoder.}
For the initial positions of the compact vectors $\mathcal{F}$, we follow VecSet \cite{zhang2023vecset} and adopt input-dependent positions, sampled from the input point cloud.
We note that our method can also incorporate learnable positional embeddings, similar to VecSet.
To encode point features, as well as the positions of point patches and compact vectors, we utilize a shared positional embedding function.
The positional embedding function follows the design used in VecSet, which comprises learnable weights and a linear layer.
When processing point patches with self-attention layers, we introduce an additional \texttt{CLS} token, following \cite{yu2024titok}, to aggregate global information.
After the last encoder block, we additionally apply one cross-attention layer to further project high-resolution features into compact latent vectors.
We use the KL block design commonly used in 2D generative models, which is also employed in VecSet.

\paragraph{Decoder.}
After pruning, the remaining tokens are processed using ViT-style transformer blocks. In this stage, we aggregate the pruned tokens into 8 merged tokens via cross-attention and concatenate them with the input sequence to leverage additional information effectively.
To reconstruct the full triplanes, we first linearly project the initial triplane tokens into dense triplane features. The processed tokens are then linearly projected and scattered into the dense triplanes. Finally, we apply a sigmoid activation to the uncertainty values to compute importance scores, which are multiplied with the scattered triplane features. The final triplane features are obtained by adding these weighted features to the initial dense triplane features.

\section{Training objective}

\paragraph{First stage.}
In the first stage training, we train the autoencoder model to minimize the reconstruction loss.
Similar to VecSet, the reconstruction loss computed as binary cross entropy between the final occupancy predictions and the ground truth:
\begin{equation}
\begin{aligned}
    \tilde{\mathcal{L}}_{recon} &= \mathbb{E}_{\textbf{q} \in \mathcal{Q}_{vol}} 
    \left[\textrm{BCE}(o(\mathbf{q}), \hat{o}(\mathbf{q}))\right] \\
    &\quad + 0.1 \cdot \mathbb{E}_{\textbf{q} \in \mathcal{Q}_{near}} 
    \left[\textrm{BCE}(o(\mathbf{q}), \hat{o}(\mathbf{q}))\right],
\end{aligned}
\end{equation}
where $o(\mathbf{q})$ is the predicted occupancy using the final triplane features, and $\hat{o}(\mathbf{q})$ is the corresponding ground truth occupancy.
The training objective of the first stage can be expressed as
\begin{equation}
    \mathcal{L}_{ae} = \tilde{\mathcal{L}}_{recon} +
    \mathcal{L}_{recon} + 
    \lambda_{unc} \cdot \mathcal{L}_{unc},
\end{equation}
where $\mathcal{L}_{recon}$ is the reconstruction loss computed using the initial triplane tokens, and $\mathcal{L}_{unc}$ is the loss for training the uncertainty head.
Both losses follow the procedure described in the main paper.
To properly train the uncertainty head, we clip the initial query-wise reconstruction loss $\mathcal{L}_{recon}(\mathbf{q})$, computed using the initial triplane features, to be in a range $[0, 1]$.

\paragraph{Second stage.}
In the second stage VAE training, we freeze the autoencoder components and train the KL block and the latent decoder.
The training objective of the second stage consists of the MSE loss between the features with two regularization terms:
\begin{equation}
    \mathcal{L}_{vae} = \textrm{MSE}(n(\hat{\mathcal{F}}), n(\mathcal{F})) + 
    \mathcal{L}_{ae} + \lambda_{kl} \cdot \mathcal{L}_{kl},
\end{equation}
where $n(\cdot)$ is a layer-wise normalization \cite{ba2016layernorm} without affine operations.
This normalization can be applied since our decoder only consists of transformer layers, which normalize the input features before processing.
Note that only the KL block and the latent decoder are trained in the second stage.

\section{Experimental setup details}

\subsection{Additional implementation details}
We set the channel dimension of the transformer 512 with 8 heads in all components.
Our models and training pipelines are implemented using the PyTorch framework.
For the ShapeNet experiments, we use the AdamW optimizer with a learning rate 1e-4, weight decay 0.01, and the effective batch size 256 to train the autoencoders and 1024 to train the VAEs.
In the second stage, the learning rate is decayed by 0.5 after 960, 1120, 1280, 1440 epochs.
For the Objaverse experiments, we use the same configurations for the first stage, and we decay the learning rate after 120, 140, 160, 180 epochs.
We set $\lambda_{unc}=0.01$ and $\lambda_{kl}=0.001$.
We use the automatic mixed precision (AMP) and the flash attention \cite{dao2022flashattention} built in for the PyTorch framework, which are disable when measuring the efficiency of our method for precise evaluations.
While we do not manually align orientations of objects and do not apply rotation augmentations, our model could be improved by applying random rotation augmentations -- particularly on the Objaverse dataset, which includes objects with inconsistent orientations.

\subsection{Evaluation protocols}
In our experiments, we follow the commonly used evaluation protocols from SDF-StyleGAN~\cite{zheng2022sdf} and VecSet~\cite{zhang2023vecset} as described in the main paper.

For reconstruction experiments, we measure Chamfer Distance (CD), volumetric Intersection-over-Union (IoU) and F-score (F1) computed using the chamfer distance.
We use the threshold 0.02 to compute F1 for ShapeNet, while we slightly increase the threshold to 0.05 for Objaverse, as the Objaverse dataset comprises more complicated objects.

\begin{figure}[t]
    \centering
    \includegraphics[width=1\linewidth]{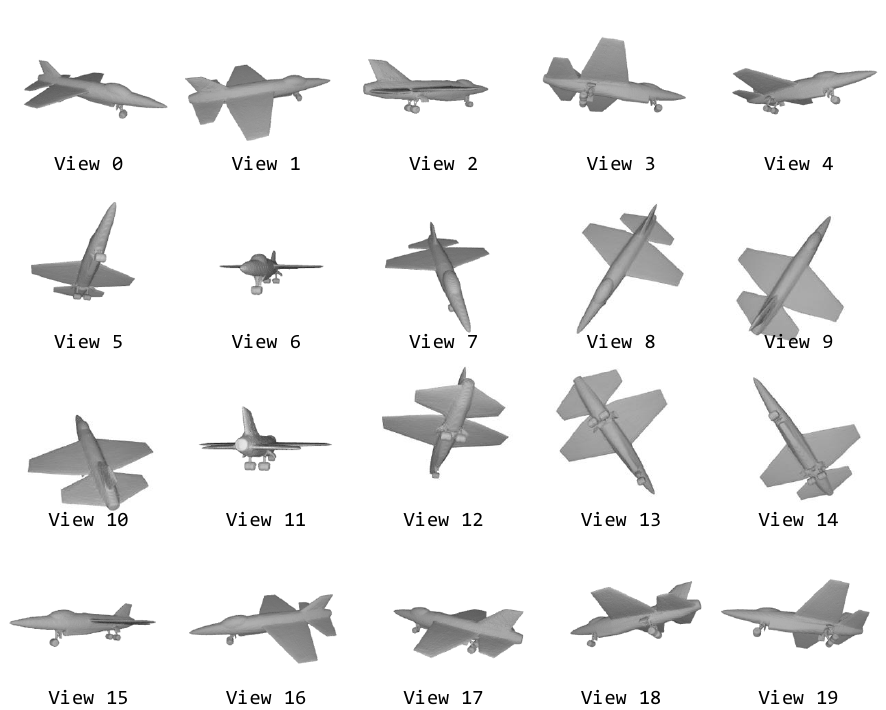}
    \vspace{-15pt}
    \caption{
    \textbf{Example visualizations of 20 rendered images}, used for computing Rendering-FID. The image resolution is $299 \times 299$. 
    }
    \label{fig:20views}
    \vspace{-10pt}
\end{figure}
To evaluate the generation performance, we employ the FID-based scores as our main metrics, which are Rendering-FID and Surface-FPD.
We note that these FID-based metrics are wide in more recent research \cite{zheng2022sdf, xiong2024octfusion, zhang2023vecset, zhang2024lagem, zhang20223dilg}, as it can better take human perception into consideration \cite{zheng2022sdf}.
For Rendering-FID, we first normalize the surface mesh to a unit sphere and then render shading images from 20 uniformly distributed views (see \Fref{fig:20views}). 
To measure Surface-FPD, we sample 4,096 points from the mesh surface and then feed them into pretrained PointNet++~\cite{qi2017pointnet++} to extract global feature vectors. 
This PointNet++ network is pretrained on ShapeNet-v2 for shape classification. using the same train/valid/test split as in our experiments.

We also measure additional metrics, MMD, COV, 1-NNA, computed using Chamfer distance (CD) and Earth Mover's distance (EMD) \cite{rubner2000earth}.
To evaluate generation performance, we use Coverage (COV), Minimum Matching Distance (MMD), 1-Nearest Neighbor Accuracy (1-NNA).
To compute these metrics, we compute the pairwise distances between the generated set $S_g$ and the reference set $S_r$.
We compute COV and MMD by using the test test as $S_r$, and generate $5|S_r|$ shapes as $S_g$.
To compute 1-NNA, we set $|S_g| = |S_r|$.
For these metrics, we sample 2,048 points from the surface of each mesh.

For category-conditioned generation, we generate 2,000 objects per category for evaluation. 
The test set distribution is as follows: airplane (202), car (175), chair (338), table (421), rifle (118).
For COV and MMD, we sample from the generated objects per category: airplane (1,010), car (875), chair (1,690), table (2,000), and rifle (590). 
For Rendering-FID and Surface-FPD, 2000 objects are sampled from the training set and compared with the generated objects.


To evaluate efficiency, we measure throughput on a single A6000 GPU. The batch size is adjusted to maximize GPU utilization, based on the model with the highest GPU memory consumption.

\section{Additional experiments and results}

\begin{table}[]
\setlength{\tabcolsep}{4pt}
\centering
\footnotesize
\begin{tabular}{cc|ccc}
\toprule
Method & Freeze & IoU (\%)$\uparrow$ & CD$\downarrow$ & F1 (\%)$\uparrow$ \\
\hline
PerceiverIO (FPS query) & \checkmark & 91.7 & 0.018 & 93.4  \\
PerceiverIO (learnable query) & \checkmark & 92.5 & 0.017 & 94.6  \\
PerceiverIO (learnable query) & & 94.0 & 0.015 & 96.1  \\
Ours ($M=32$) & & 96.1 & 0.012 & 98.0 \\
\bottomrule
\end{tabular}

\vspace{-5pt}
\caption{
\textbf{Results of PerceiverIO} with $M=32$ and pretrained VecSet ($M=512$) on ShapeNet.
We also report variants with frozen VecSet, and with learnable queries replaced by FPS as used in VecSet.
Results of autoencoders are reported.
}
\vspace{-5pt}
\label{table:comp_perceiver}
\end{table}
\subsection{Comparison with PerceiverIO}
To validate the compression capability of our method, we compare it with PerceiverIO \cite{jaegle2021perceiver}, which is trained to compress the original latent space of VecSet.
Specifically, PerceiverIO takes the latent vectors from the decoder of a pretrained VecSet as input and outputs compressed latent vectors.
The remaining architecture of this model (\ie, decoder, neural fields decoding) follows the VecSet pipeline.
We also evaluate several variants: one with the frozen VecSet, and another that uses FPS-based queries instead of learnable queries, following the design of VecSet.

As shown in \Tref{table:comp_perceiver}, our model outperforms several variants of PerceiverIO by a large margin.
These results indicate that existing latent compression methods, primarily explored in the 2D computer vision and NLP domains, require careful considerations in architecture design to effectively process 3D modality.

\begin{table}[]
\setlength{\tabcolsep}{4pt}
\centering
\footnotesize
\begin{tabular}{cccc|c}
\toprule
VecSet & $M=32$ & $M=64$ & $M=512$ & Ours ($M=64$) \\
\hline
Surface-IoU (\%)$\uparrow$ & 72.7 & 78.0 & 87.8 & 88.0 \\
\bottomrule
\end{tabular}

\setlength{\tabcolsep}{12pt}
\vspace{1pt}
\begin{tabular}{ccccc}
\toprule
Pruning ratio & 0\% & 50\% & 75\% & 90\% \\
\hline
Surface-IoU (\%)$\uparrow$ & 87.9 & 87.8 & 87.6 & 87.1 \\
\bottomrule
\end{tabular}
\vspace{-5pt}
\caption{
\textbf{Near-surface reconstruction results on ShapeNet.} To better assess the quality of reconstructed surfaces, we measure IoU using 50K points sampled near object surfaces.
\textbf{(top)} VAE comparisons with VecSet. \textbf{(bottom)} Ablation results on the pruning ratio, obtained using autoencoders with $M=32$.
}
\vspace{-5pt}
\label{table:comp_surface}
\end{table}
\subsection{Near-surface reconstruction results}
We additionally provide near-surface reconstruction IoU (Surface IoU) to better assess the quality of reconstructed surfaces.
As reported in \Tref{table:comp_surface}, our VAE with $M=64$ achieves comparable performance in near-surface regions to VecSet with $M=512$.
In addition, while pruning up to 75\% results in only a minor drop, we observe a more noticeable degradation in shape details beyond this ratio.
Based on this trade-off, we set the pruning ratio to 75\% to balance efficiency and quality (see Tab.~6 of the main paper for efficiency gains).

\begin{figure}[t]
    \begin{center}
        \includegraphics[width=0.9\linewidth]{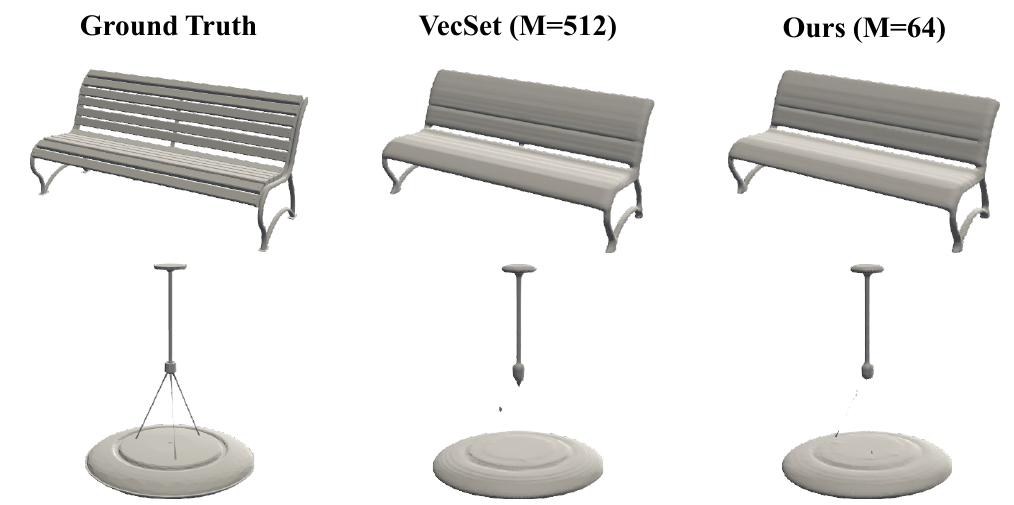}
    \end{center}
    \vspace{-10pt}
    \caption{\textbf{Failure cases} of our model and VecSet ($M=512$). We report the results of the autoencoders.}
    \label{fig:failure}
    \vspace{-5pt}
\end{figure}
\subsection{Failure cases}
Since our model employs a training strategy similar with VecSet, it shares similar failure cases with VecSet using 512 latent vectors.
As presented in \Fref{fig:failure}, both our model and VecSet struggles to model extremely thin structures of the objects, often producing over-smooth surfaces.
This can be addressed by improving the points sampling strategy for both query points and point patches, such as sharp edge sampling proposed in \cite{chen2024dora}.

\begin{figure*}[t]
    \centering
    \includegraphics[width=1\linewidth]{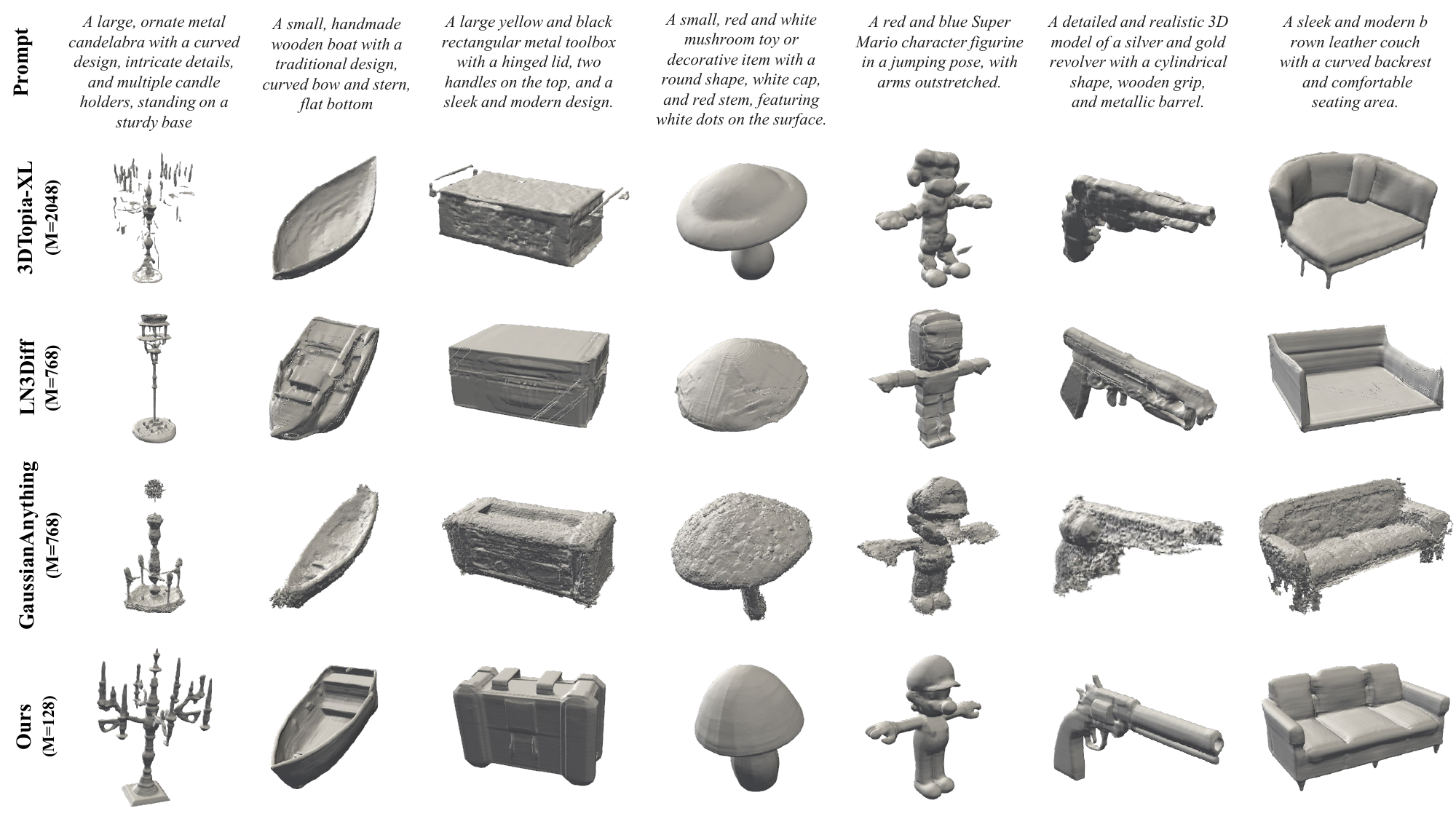}
    \vspace{-10pt}
    \caption{
    \textbf{Text-conditioned generation results.} Results of competing methods are obtained using their official source codes and pretrained weights. We report mesh outputs of these methods without textures and colors.
    }
    \label{fig:gen_text_cond}
    \vspace{-10pt}
\end{figure*}
\subsection{Text-conditioned generation}
We further evaluate the generation performance of our model on text-conditioned 3D object generation.
To enhance quality, we train the VAE with 128 latent vectors, incorporate two additional layers before the uncertainty-guided pruning module in the decoder, and add a convolutional refinement layer after triplane reconstruction.
We also filter out low-quality objects (those with too few occupied points) from the training dataset.
Following \cite{yushi2025ga}, we use the captions provided in \cite{chen20253dtopia} and extract text embeddings using CLIP.
The diffusion model is trained on approximately 20K objects, and evaluation is performed on unseen captions randomly selected from the validation set.

As shown in \Fref{fig:gen_text_cond}, our model generates high-quality objects using only 128 latent vectors, while existing methods typically require significantly more.
Note that the compared methods are trained with different dataset sizes and are designed to generate colored meshes, making direct comparisons difficult.
Our model may also benefit from larger-scale training to improve prompt-following ability.
Nevertheless, we highlight that our approach can achieve high-quality generation with greater efficiency.

\begin{table}[]
\setlength{\tabcolsep}{6pt}
\centering
\footnotesize
\begin{tabular}{c|ccc}
\toprule
 & IoU (\%)$\uparrow$ & CD$\downarrow$ & F1 (\%)$\uparrow$ \\
\hline
Ours ($M=16$) & 95.6 & 0.013 & 97.8 \\
Ours ($M=32$) & 96.1 & 0.012 & 98.0 \\
Ours ($M=64$) & 96.5 & 0.012 & 98.2 \\
Ours ($M=128$) & 96.7 & 0.012 & 98.2 \\
\hline
Learnable positions &  96.0 & 0.012 & 98.0 \\
Input-dependent positions & 96.1 & 0.012 & 98.0 \\
\hline
Single-stage training & 94.4 & 0.014 & 97.1 \\
Two-stage training & 96.1 & 0.012 & 98.0 \\
\hline
Confidence \cite{wang2024dust3r} based uncertainty & 95.3 & 0.015 & 97.1 \\
Recon. based uncertainty & 96.1 & 0.012 & 98.0 \\
\bottomrule
\end{tabular}
\vspace{-5pt}
\caption{
\textbf{Additional ablation results on ShapeNet}. Top-to-bottom: ablation results on the number of latent vectors, results with the learnable latent positions, results without two-stage training, and results with the confidence-based training objective for the uncertainty head.
}
\label{table:supp_ablation}
\vspace{-5pt}
\end{table}

\subsection{Additional ablation study}
We further explore the behaviors of our method through the additional ablation studies.
Consistent with the main paper, all ablations are conducted using the autoencoder with $M=32$ on ShapeNet \cite{chang2015shapenet}.

\paragraph{The number of latent vectors.}
As shown in \Tref{table:supp_ablation} (first group),  the performance of our method improves as the number of latent vectors increases.
However, the performance gain diminishes beyond a certain number of latent vectors.
This suggests that most of the essential information can be effectively encoded with $M=32$ or $M=64$ latent vectors, while additional vectors contribute only marginal improvements beyond this point.

\paragraph{Positions of the latent vectors.}
Our method can also employ learnable positional embeddings for the latent vectors.
As reported in \Tref{table:supp_ablation} (second group), the model with learnable latent positions achieves similar reconstruction quality with the model with input-dependent positions.
Since the learnable latent positions can provide orders of the latent vectors, they can also be a useful option for the cases where we need to model the latent vectors as a ordered sequence, \eg, autoregressive generation.

\paragraph{Two-stage training.}
We evaluate the two-stage training scheme as presented in \Tref{table:supp_ablation} (third group).
The two-stage training effectively improves the overall performance.
In contrast, single-stage training makes autoencoders learn to compress both the number of latent vectors and along the channel dimension.
Additionally, training uncertainty head with an auxiliary loss further complicates the training.
These complicated training objectives lead to a noticeable performance drop.
Therefore, we separate the autoencoder training from the training of channel compression modules.

\paragraph{Uncertainty head objective.}
Finally, we replace the training objective of the uncertainty head with the confidence-based objective used in \cite{wang2024dust3r}.
As shown in \Tref{table:supp_ablation} (fourth group), the model trained to predict reconstruction errors achieves better performance.
We attribute this to the fact that training the uncertainty head with a more explicit objective--namely, estimating the reconstruction error of each triplane token--is more effective than using a confidence-based objective for token pruning.

\begin{table*}[]
    \setlength{\tabcolsep}{6pt}
    \centering
    \footnotesize
    \begin{tabular}{c|r|ccccc}
    \toprule
                                    & Method            & Airplane	& Car	    & Chair	    & Table	    & Rifle \\ \hline
    \multirow{7}{*}{MMD-CD}	        & 3DILG	            & 3.702	    & 4.353	    & 9.243	    & 10.526	& 3.529 \\
                                    & GEM3D	            & 3.587	    & 4.160	    & 8.680	    & 7.652	    & 2.828 \\
                                    & VecSet (M=32)	    & 3.097	    & 4.155	    & 8.173	    & 6.883	    & 2.689 \\
                                    & VecSet (M=64)	    & 3.121	    & 4.173	    & 8.360	    & 6.955	    & 2.841 \\
                                    & VecSet (M=512)	& 3.059	    & 3.921	    & 7.821	    & 6.527	    & 2.707 \\
                                    & Ours (M=32)	    & 3.275	    & 4.054	    & 8.067	    & 6.885	    & 2.820 \\
                                    & Ours (M=64)	    & 3.240	    & 3.971	    & 8.465	    & 6.695	    & 2.755 \\ \hline
    \multirow{7}{*}{MMD-EMD}	    & 3DILG	            & 9.04	    & 10.28	    & 13.35	    & 13.46	    & 8.78  \\
                                    & GEM3D	            & 8.75	    & 9.62	    & 13.12	    & 11.91	    & 8.67  \\
                                    & VecSet (M=32)	    & 8.88	    & 10.18	    & 13.12	    & 11.71	    & 8.29  \\
                                    & VecSet (M=64)	    & 8.82	    & 10.02	    & 13.04	    & 11.84	    & 8.71  \\
                                    & VecSet (M=512)	& 8.64	    & 9.78	    & 12.69	    & 11.55	    & 8.40  \\
                                    & Ours (M=32)	    & 8.79	    & 9.84	    & 13.01	    & 11.76	    & 8.52  \\
                                    & Ours (M=64)	    & 8.71	    & 9.83	    & 13.09	    & 11.59	    & 8.66  \\ \hline
    \multirow{7}{*}{COV-CD}	        & 3DILG	            & 67.33	    & 41.71	    & 70.41	    & 46.08	    & 65.25 \\
                                    & GEM3D	            & 74.26	    & 49.71	    & 68.93	    & 75.06	    & 65.25 \\
                                    & VecSet (M=32)	    & 88.12	    & 74.29	    & 88.76	    & 90.02	    & 81.36 \\
                                    & VecSet (M=64)	    & 86.14	    & 76.57	    & 85.50	    & 88.12	    & 88.14 \\
                                    & VecSet (M=512)	& 85.15	    & 76.00	    & 87.28	    & 88.60	    & 88.98 \\
                                    & Ours (M=32)	    & 84.65	    & 77.71	    & 87.28	    & 87.89	    & 87.29 \\
                                    & Ours (M=64)	    & 84.16	    & 79.43	    & 84.32	    & 90.74	    & 91.53 \\ \hline
    \multirow{7}{*}{COV-EMD}	    & 3DILG	            & 72.77	    & 41.71	    & 76.33	    & 52.02	    & 71.19 \\
                                    & GEM3D	            & 73.27	    & 58.86	    & 64.2	    & 74.58	    & 59.32 \\
                                    & VecSet (M=32)	    & 85.64	    & 60.57	    & 89.05	    & 91.69	    & 87.29 \\
                                    & VecSet (M=64)	    & 90.1	    & 65.71	    & 84.91	    & 91.69	    & 89.83 \\
                                    & VecSet (M=512)	& 88.61	    & 63.43	    & 85.8	    & 89.55	    & 91.53 \\
                                    & Ours (M=32)	    & 86.14	    & 69.71	    & 89.05	    & 89.79	    & 86.44 \\
                                    & Ours (M=64)	    & 89.6	    & 70.29	    & 86.69	    & 92.16	    & 88.14 \\ \hline
    \multirow{7}{*}{1-NNA-CD}	    & 3DILG	            & 61.39	    & 61.43	    & 57.69	    & 68.88	    & 58.05 \\
                                    & GEM3D	            & 54.46	    & 58.00	    & 53.55	    & 53.44	    & 54.24 \\
                                    & VecSet (M=32)	    & 54.21	    & 64.29	    & 53.85	    & 50.95	    & 48.31 \\
                                    & VecSet (M=64)	    & 52.97	    & 63.71	    & 54.29	    & 52.97	    & 51.27 \\
                                    & VecSet (M=512)	& 51.98	    & 62.00	    & 54.73	    & 50.59	    & 49.58 \\
                                    & Ours (M=32)	    & 53.71	    & 60.00	    & 52.07	    & 50.59	    & 49.15 \\
                                    & Ours (M=64)	    & 53.22	    & 61.43	    & 53.99	    & 52.02	    & 53.81 \\ \hline
    \multirow{7}{*}{1-NNA-EMD}	    & 3DILG	            & 55.69	    & 60.57	    & 58.73	    & 66.86	    & 57.63 \\
                                    & GEM3D	            & 52.72	    & 59.43	    & 56.21	    & 56.65	    & 55.08 \\
                                    & VecSet (M=32)	    & 56.93	    & 63.14	    & 52.96	    & 52.02	    & 48.73 \\
                                    & VecSet (M=64)	    & 51.98	    & 62.29	    & 52.96	    & 53.68	    & 53.81 \\
                                    & VecSet (M=512)	& 53.96	    & 61.71	    & 53.7	    & 52.14	    & 53.81 \\
                                    & Ours (M=32)	    & 52.72	    & 57.71	    & 53.11	    & 53.09	    & 53.81 \\
                                    & Ours (M=64)	    & 53.96	    & 60.57	    & 52.22	    & 51.07	    & 52.54 \\ \hline
    \multirow{7}{*}{Rendering-FID}	& 3DILG	            & 40.49	    & 134.00	& 38.20	    & 62.63	    & 48.81 \\
                                    & GEM3D	            & 34.97	    & 108.86	& 32.24	    & 31.95	    & 31.89 \\
                                    & VecSet (M=32)	    & 60.25	    & 132.91	& 56.26	    & 39.60	    & 38.76 \\
                                    & VecSet (M=64)	    & 52.75	    & 107.52	& 47.96	    & 34.77	    & 29.35 \\
                                    & VecSet (M=512)	& 31.57	    & 108.41	& 26.31	    & 31.78	    & 22.79 \\
                                    & Ours (M=32)	    & 33.57	    & 79.91	    & 27.65	    & 26.89	    & 18.67 \\
                                    & Ours (M=64)	    & 31.72	    & 79.97	    & 27.08	    & 27.53	    & 18.95 \\ \hline
    \multirow{7}{*}{Surface-FPD}	& 3DILG	            & 1.13	    & 4.82	    & 1.45	    & 2.55	    & 2.97  \\
                                    & GEM3D	            & 0.82	    & 1.05	    & 0.99	    & 1.02	    & 1.13  \\
                                    & VecSet (M=32)	    & 0.39	    & 2.51	    & 0.39	    & 0.29	    & 0.42  \\
                                    & VecSet (M=64)	    & 0.27	    & 1.87	    & 0.41	    & 0.29	    & 0.31  \\
                                    & VecSet (M=512)	& 0.26	    & 1.27	    & 0.36	    & 0.31	    & 0.41  \\
                                    & Ours (M=32)	    & 0.26	    & 1.28	    & 0.33	    & 0.24	    & 0.25  \\
                                    & Ours (M=64)	    & 0.25	    & 1.21	    & 0.34	    & 0.26	    & 0.24  \\
    \bottomrule
    
    \end{tabular}
    \vspace{-5pt}
    \caption{
    \textbf{Class-conditioned generation results on ShapeNet.}
    We report per-category evaluation results.
    The scales of MMD are $10^{-3}$, and $10^{-2}$ for CD, and EMD, respectively.
    }
    \label{table:supp_class_cond}
    \vspace{-10pt}
\end{table*}
\begin{figure*}[t]
    \centering
    \includegraphics[width=1\linewidth]{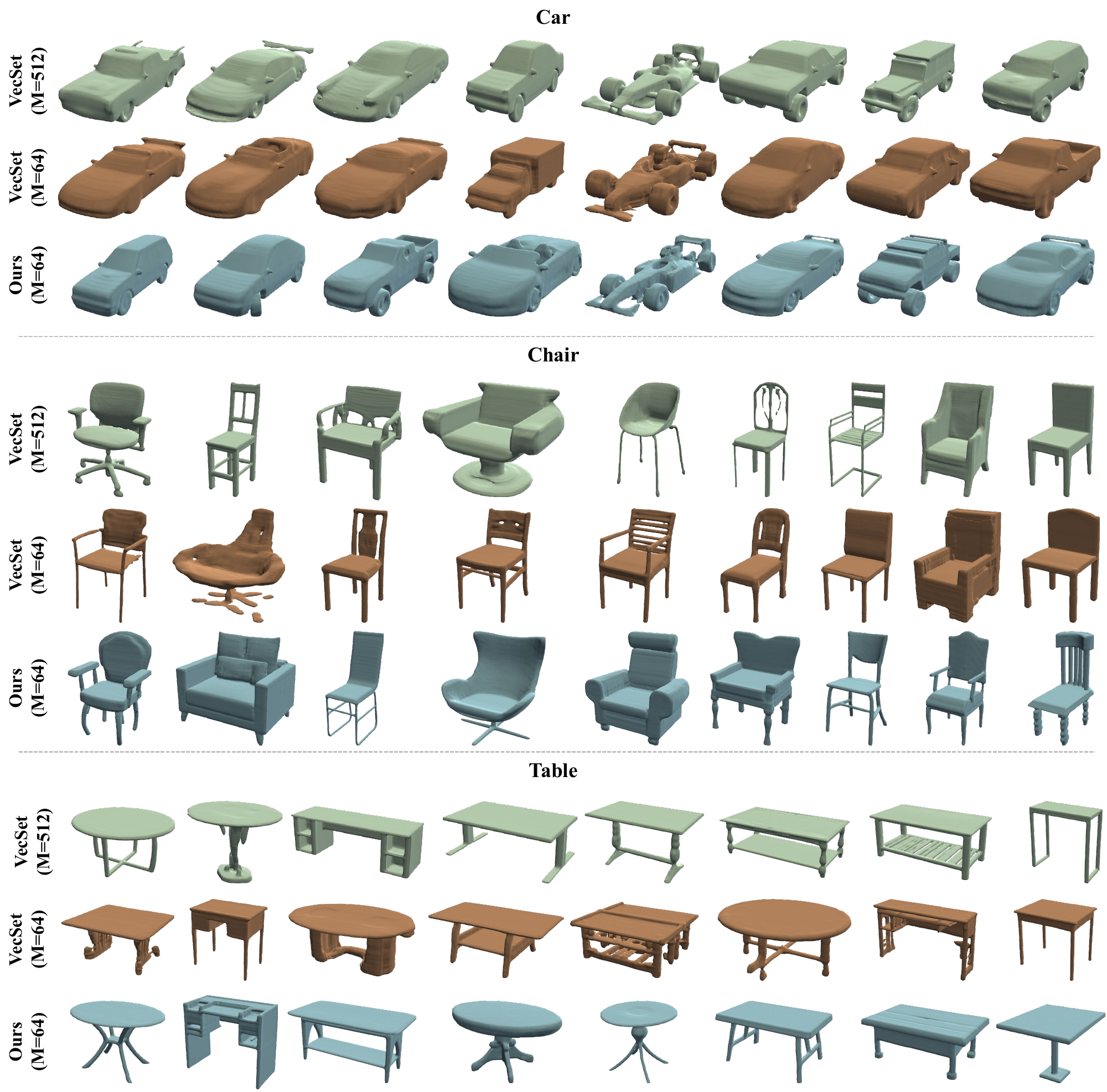}
    \vspace{-10pt}
    \caption{
    \textbf{Additional class-conditioned generation results.} We present the generated results of \textit{car}, \textit{chair}, and \textit{table}.
    }
    \label{fig:supp_gen_cond_class3}
\end{figure*}
\begin{figure*}[t]
    \centering
    \includegraphics[width=1\linewidth]{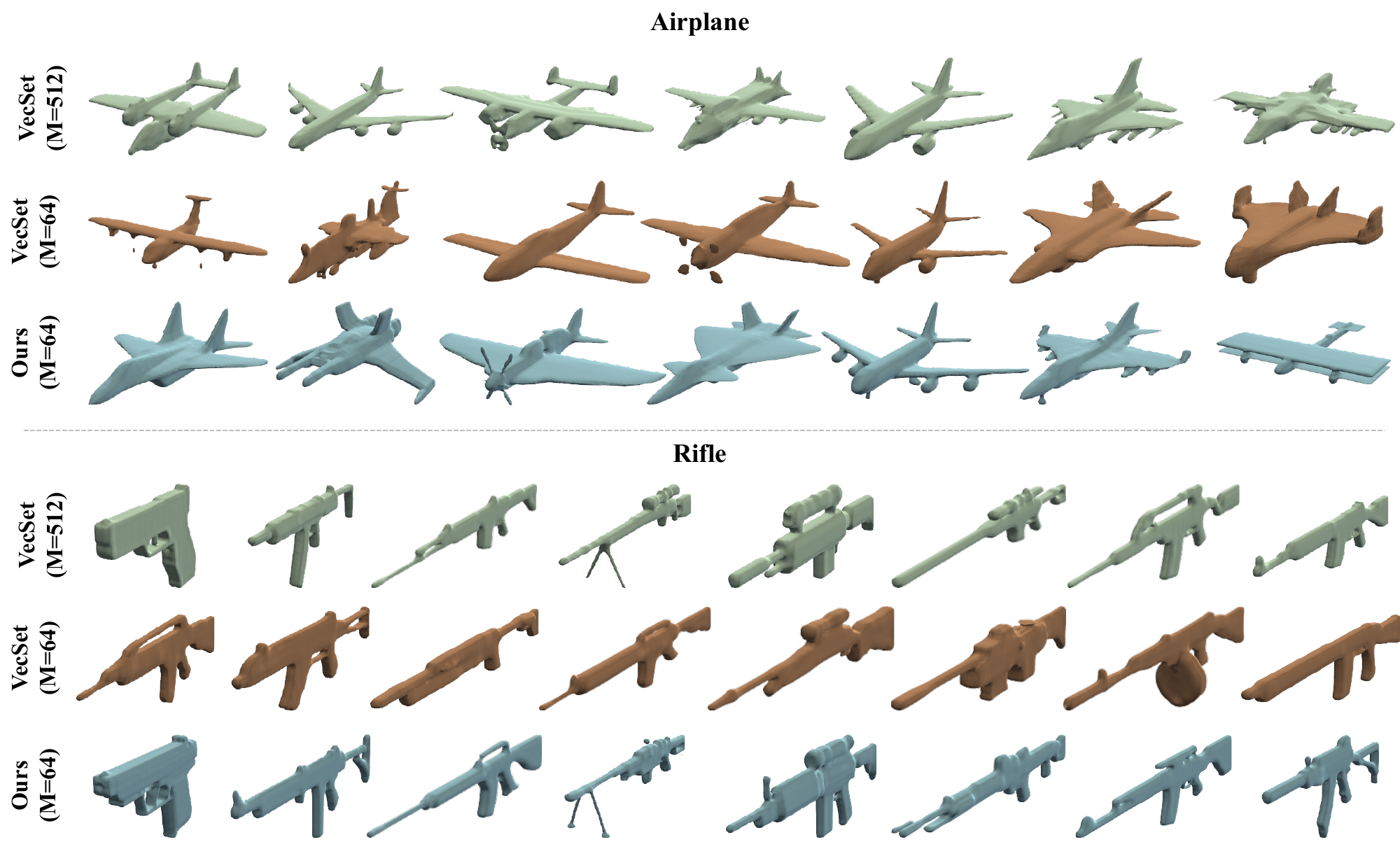}
    \vspace{-10pt}
    \caption{
    \textbf{Additional class-conditioned generation results.} We present the generated results of \textit{airplane} and \textit{rifle}.
    }
    \label{fig:supp_gen_cond_class2}
\end{figure*}
\subsection{Class-conditioned generation}
We provide per-category evaluation results of the category-conditioned generation in \Tref{table:supp_class_cond}.
We also present additional qualitative generation results of all 5 categories in Figures~\ref{fig:supp_gen_cond_class3} and \ref{fig:supp_gen_cond_class2}.

\begin{figure*}[t]
    \centering
    \includegraphics[width=1\linewidth]{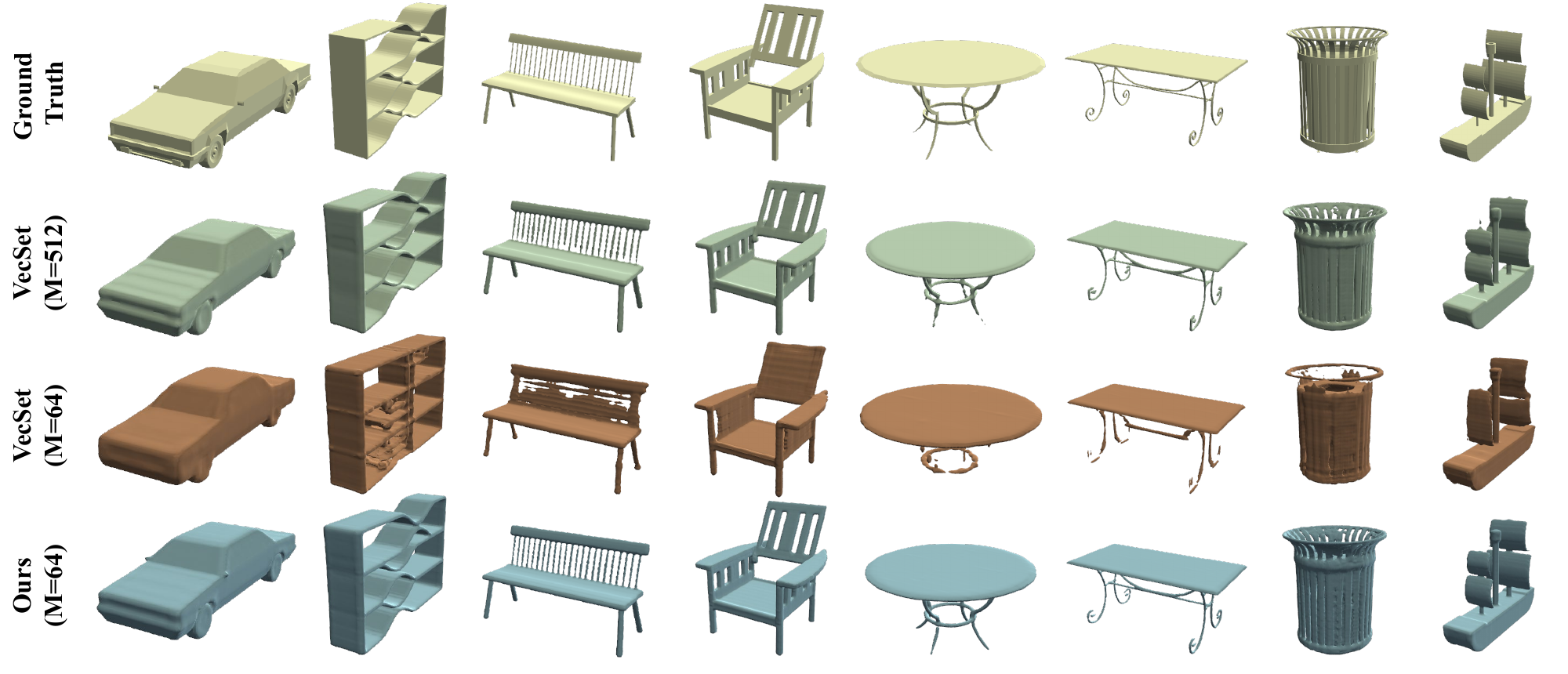}
    \vspace{-10pt}
    \caption{
    \textbf{Additional reconstruction results on ShapeNet.} We present the reconstruction results of VAEs.
    }
    \label{fig:supp_recon_shapenet}
    \vspace{-10pt}
\end{figure*}
\begin{figure*}[t]
    \centering
    \includegraphics[width=1\linewidth]{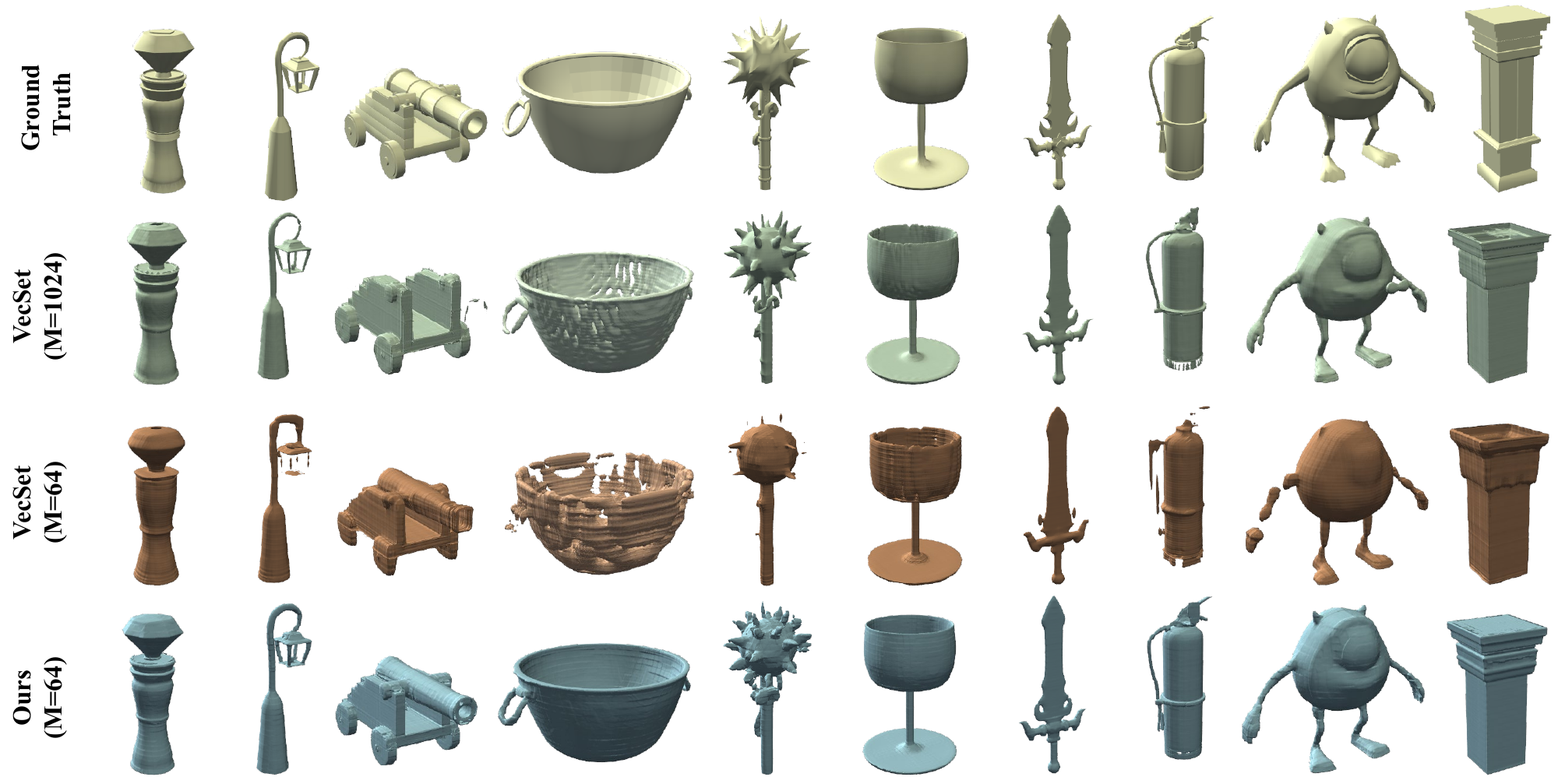}
    \vspace{-10pt}
    \caption{
    \textbf{Additional reconstruction results on Objaverse.} We present the reconstruction results of VAEs.
    }
    \label{fig:supp_recon_objaverse}
    \vspace{-10pt}
\end{figure*}
\subsection{Additional reconstruction results}
Finally, we provide additional qualitative reconstruction results. \Fref{fig:supp_recon_shapenet} presents reconstruction results on ShapeNet, and \Fref{fig:supp_recon_objaverse} presents reconstruction results on Objaverse.


\end{document}